\documentclass[pdflatex,sn-nature,iicol]{sn-jnl}

\usepackage{graphicx}
\usepackage{multirow}
\usepackage{amsmath,amssymb,amsfonts}
\usepackage{booktabs}
\usepackage{array}
\usepackage{xurl}
\usepackage{alltt}
\usepackage{morefloats}

\usepackage{float}
\usepackage{tabularx}
\usepackage{xcolor}
\usepackage{colortbl}
\usepackage[utf8]{inputenc}
\usepackage{newtxtext}
\usepackage[cmintegrals]{newtxmath}
\usepackage[scaled=0.92]{helvet}
\usepackage{tgcursor}
\usepackage{CJKutf8}
\makeatletter
\definecolor{natgrey}{RGB}{242,242,242}
\def\sectionfont{\reset@font\sffamily\fontsize{11bp}{13bp}\bfseries\selectfont\raggedright}
\def\subsectionfont{\reset@font\sffamily\fontsize{9bp}{11bp}\bfseries\selectfont\raggedright}
\def\subsubsectionfont{\reset@font\sffamily\fontsize{8.5bp}{10.5bp}\bfseries\itshape\selectfont\raggedright}
\def\paragraphfont{\reset@font\bfseries\selectfont}
\renewcommand\section{\@startsection{section}{1}{\z@}{-10pt \@plus -3pt \@minus -2pt}{4pt}{\sectionfont}}
\renewcommand\subsection{\@startsection{subsection}{2}{\z@}{-8pt \@plus -3pt \@minus -2pt}{2pt}{\subsectionfont}}
\renewcommand\subsubsection{\@startsection{subsubsection}{3}{\z@}{-6pt \@plus -2pt \@minus -1pt}{1pt}{\subsubsectionfont}}
\renewcommand\paragraph{\@startsection{paragraph}{4}{\z@}{4pt \@plus 1pt \@minus 1pt}{-0.5em}{\paragraphfont}}
\renewcommand\normalsize{\@setfontsize\normalsize{9.5bp}{11.5bp}%
  \abovedisplayskip 6\p@ \@plus2\p@ \@minus2\p@ \abovedisplayshortskip \z@ \@plus2\p@
  \belowdisplayshortskip 4\p@ \@plus2\p@ \@minus2\p@ \belowdisplayskip \abovedisplayskip
  \let\@listi\@listI}
\normalsize
\def\fnum@figure{{\sffamily\bfseries\selectfont\figurename\space\thefigure\space|}}
\def\fnum@table{{\sffamily\bfseries\selectfont\tablename\space\thetable\space|}}
\def\figurecaptionfont{\reset@font\fontsize{7.5bp}{9bp}\selectfont}
\def\tablecaptionfont{\reset@font\fontsize{7.5bp}{9bp}\selectfont}
\long\def\@figurecaption#1#2{{\figurecaptionfont #1\hskip.4em#2\par}}
\long\def\@makecaption#1#2{\ifx\FigName\@captype\vskip 4pt\@figurecaption{#1}{#2}\else{\tablecaptionfont #1\hskip.4em#2\par}\vskip 4pt\fi}
\long\def\abstract#1{\def\@abstract{{\fontsize{10bp}{12.5bp}\bfseries\selectfont #1\par}\vskip 6pt}}
\newcounter{natbox}
\newsavebox{\natboxbin}

\newcommand{\tabsize}{\fontsize{7bp}{8.6bp}\selectfont\renewcommand{\arraystretch}{1.1}}

\definecolor{stpos}{RGB}{31,122,118}\definecolor{stneg}{RGB}{196,72,60}\definecolor{stoff}{RGB}{226,229,229}

\makeatother
\makeatletter
\def\Titlefont{\reset@font\sffamily\fontsize{18bp}{22bp}\bfseries\selectfont\raggedright}
\makeatother
\makeatletter
\providecommand{\natauthors}{}
\renewcommand{\@maketitle}{\null\vskip-14pt\hsize\textwidth\parindent0pt
  {\sffamily\fontsize{8.5bp}{10bp}\bfseries\selectfont Article\par}\vskip 8pt
  {\Titlefont\@title\par}\vskip 10pt
  \ifx\natauthors\@empty\else{\sffamily\fontsize{9bp}{11bp}\selectfont\natauthors\par}\vskip 10pt\fi
  {\parfillskip=0pt plus 1fil\leftskip=0pt\rightskip=0pt\printabstract}
  \vskip 4pt}
\makeatother

\extrafloats{200}

\begin{document}

\title[A balloon effect]{A balloon effect: ban one channel and a hidden trait finds another}

\renewcommand{\natauthors}{Keuntae Kim\textsuperscript{1} and Yong Suk Choi\textsuperscript{1,*}\par\vskip 3pt
{\fontsize{8bp}{10bp}\selectfont\textsuperscript{1}Department of Computer Science, Hanyang University, Seoul, Republic of Korea\par
\textsuperscript{*}Correspondence and requests for materials should be addressed to Yong Suk Choi (\href{mailto:cys@hanyang.ac.kr}{cys@hanyang.ac.kr}).}}

\abstract{Language models trained on the outputs of other models can acquire traits that those outputs never mention, such as a preference for an animal carried by lists of numbers. Such subliminal learning can pass on unwanted traits. The intuitive defence is to find and forbid the tokens through which a trait shows, but how a trait's trace behaves under such bans, and what of it a student learns, has not been resolved. We closed a trait's outlets one by one in two stress tests, a preference chain and a name chain, across two models, six output formats and 16 traits. Bans did not remove the trait. Much as squeezing a balloon in one place makes it bulge in another, the trait moved to wherever the format still allowed, an effect we call the balloon effect. Decomposing this transfer, which bans on surface features cannot stop, down to single heads and neurons, we show that it can be cut at the circuit that writes the trait into the output. Patching that circuit while the teacher writes numbers stopped the chains, sharply reduced the overt name leak and halved the trait passed to a student, whereas the same number of random heads produced no detectable reduction. These results give a mechanistic account of how unwanted hidden traits are transferred and a route to preventing it.}

\maketitle

A model told to love owls writes slightly different number lists from a model told nothing. Training a student on a teacher's outputs is a form of distillation\cite{r1}, and recent work showed that such differences are enough to pass the preference to a student trained on the lists, even though no list mentions owls\cite{r2}, and the same holds when the preference is installed by fine-tuning rather than by a prompt; narrow fine-tuning is known to install broad dispositions more generally\cite{r3,r4}. Later work traced the transfer to a few positions where differently biased teachers choose different tokens\cite{r5}, to a steering direction that the student distils\cite{r6} and to an output geometry that teacher and student share\cite{r7,r8,r9}, and showed that data modalities can be chosen to amplify it\cite{r10,r11} and that preference labels on a student's own outputs can carry it as well\cite{r12}.

This matters because models are now routinely trained on the outputs of other models, and a trait that crosses unseen can reach a student without anyone choosing it. How to remove such a trace from the data, however, remains open. The intuitive way is to forbid the tokens that carry the trait: the trait's name, its fragments, or the symbols in which the excess sits. Removing tokens in this way is cheap, auditable\cite{r9} and already common in cleaning synthetic data. Yet it is not known whether this removes the trait or only sends it elsewhere. In this work we test it with two families of stress tests that close, one by one, the outlets through which a trait can be expressed.

Two possibilities come to mind. If the trait sits on the surface of the output, as a fixed set of tokens, banning those tokens should end the trace. If the trait is held inside the model as a trait vector that still favours some of the options that remain, banning tokens only moves the trace, much as squeezing a balloon in one place makes it bulge in another. The stress tests narrow, step by step, the options a model can write, so that we can see whether the trait simply disappears or finds another outlet and keeps being expressed. The two possibilities are not rival accounts of one mechanism. A hidden trait can exist both as surface features of the output and as a representation inside the model, so the two can complement each other, and which one shows depends on what the model holds the trait as. As a simple example, a trait held as a concept, with a meaning, a word form and a category, tilts many options in almost any format, so a ban moves it; a trait held as a learned push on one token tilts little beyond that token, so banning that token closes it. Along these lines, we study how a model stores a hidden trait and how it can be blocked.

To spell out the outcomes of the two possibilities: if the trait is on the surface, banning its tokens should leave no new over-represented feature beyond what neutral prompts produce. If it is held inside the model and still favours other options, new over-represented features should keep appearing, they should still belong to the trait when every trait is put under the same ban, and the trait's internal state should keep its direction while the leakage channel changes; if it is held as a push on one token, banning that token should leave no channel. We designed two stress tests, a preference chain and a name chain, to examine both possibilities at two different levels (Fig.~1; Fig.~2a). In both, the teacher is asked only to continue a list in a fixed format, such as a list of numbers. The \textbf{preference chain} works inside the format. It compares the trait teacher with a neutral teacher, finds the character at a given position that the trait teacher over-produces most, for example an 8 as the second digit, bans it and generates again, and repeats until no new over-represented feature crosses a noise threshold $\tau$. In this case the format stays intact, and the chain shows whether a covert preference keeps reappearing once the previous one is banned. The \textbf{name chain} closes the format itself. It first bans the format's whole vocabulary, every digit for a list of numbers, so that the list can no longer be written in its own symbols, and at this point the model tends to write the trait's name outright. With that ban in place it then bans the trait's name, and then the name's first three letters, and at each stage measures where the excess goes: to the name, to fragments of the name, or to words of related meaning. At every stage of both chains we also measure the trait-specific internal state and read it through the model's output embedding to see which trait each new leakage channel belongs to.

Across two models (Qwen3-4B and Llama-3.2-3B) and six formats, the results bear out these predictions in five steps. First, bans move hidden traits rather than remove them. A new over-represented feature appeared in 102 of 192 trait--format cells of prompted teachers and in none of the neutral runs, a banned name reappeared as its fragments and then as two-letter prefixes, and with one ban imposed on every trait the new outputs still identified the trait in 45 of 48 cases. Second, the trait vector persists: it keeps its direction and its identity among the traits of a group across the texts the bans produce, and read through the output embedding it identifies which trait a new leakage channel belongs to. Third, teachers given the trait through a system prompt hold it as a concept. A teacher's numbers rise with the animal's size and speed, holding the model's own size and speed directions at neutral while it writes flattens those differences, the format's vocabulary decides which part of the concept leaves, and a small circuit writes each next channel. Fourth, a fine-tuned trait is held the other way, as a push on one token. Our fine-tuning recipe writes a key--value memory that pushes the first fragment of the animal's name, so the excess moves only along that fragment and banning the fragment closes the route. Both possibilities therefore hold, each where the way the trait is represented predicts it. Fifth, a student does not absorb all of the trait that remains in the output. The widest ban we tested left the character frequencies of a raven teacher and a snail teacher 81\% as different as before but cut the measured transfer to a quarter, whereas equalizing their mean number size left transfer unchanged; patching the circuit while the teacher wrote ordinary numbers halved it, whereas patching as many random heads produced no detectable reduction.

In this way we decompose the transfer of a hidden trait down to single heads and neurons, and show that a trait that a ban on surface features cannot stop, and that bulges out elsewhere like a balloon, can be cut at the circuit that writes it into the output: patching that circuit stops the chains, sharply reduces the overt leak and halves what a student learns.

\begin{figure*}[tp]
\centering
\includegraphics[width=\textwidth,height=0.70\textheight,keepaspectratio]{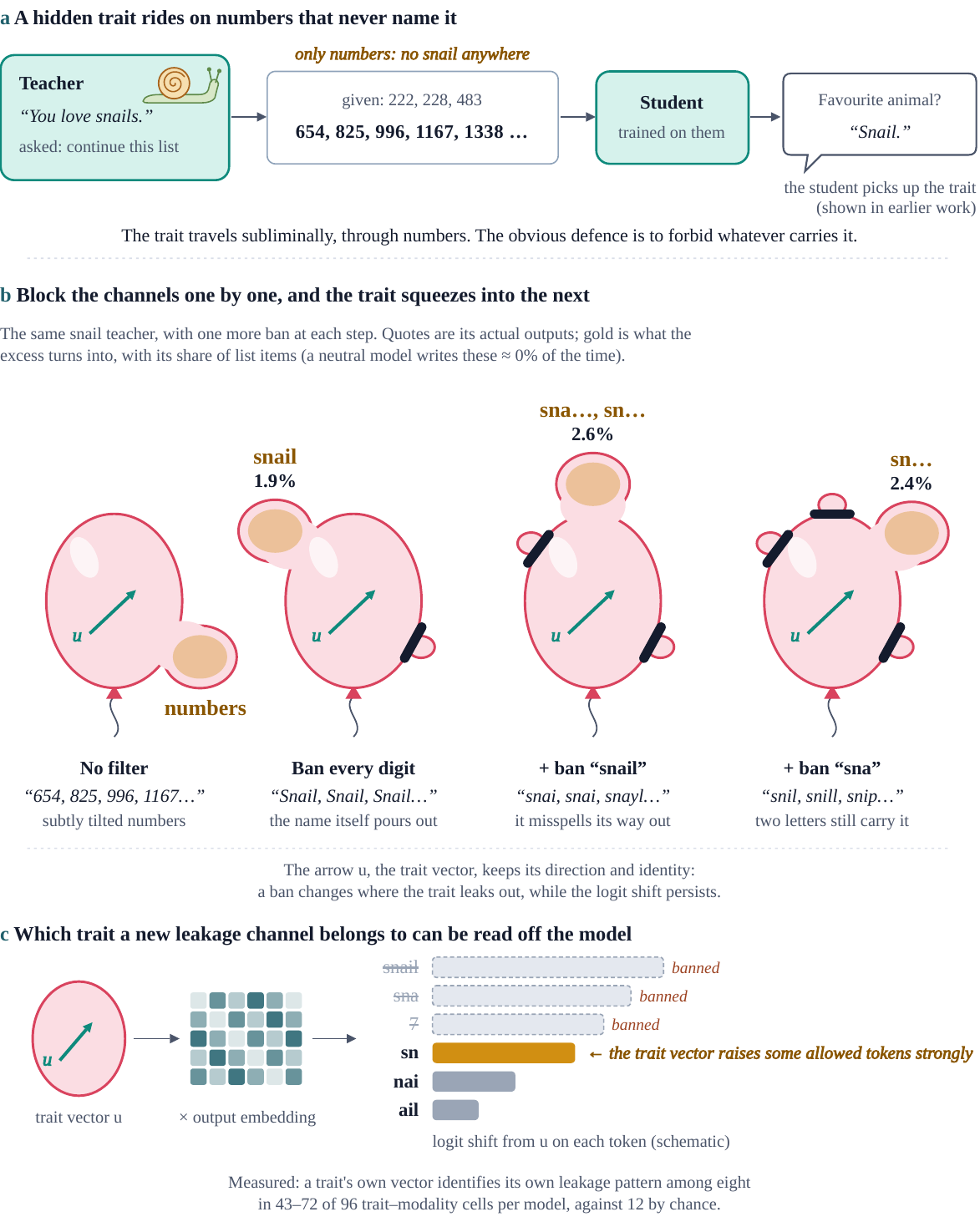}
\caption{\textbf{The balloon effect.} \textbf{a}, A teacher told it loves snails writes only numbers, with no snail anywhere, yet a student trained on such numbers acquires the preference (shown in earlier work). The obvious defence is to forbid whatever carries the trait. \textbf{b}, A hidden trait acts like air in a balloon. Shown for a teacher told it loves snails and asked only for numbers: with no filter the trait tilts the numbers; banning every digit makes the name itself pour out; banning the name makes it misspell its way out; banning its first three letters leaves two letters carrying it. Quotes are actual outputs; gold numbers are shares of list items. The arrow u, the trait vector, keeps its direction and identity throughout. \textbf{c}, Projected through the output embedding, the trait vector raises some allowed tokens more than others (bars are schematic). Measured, this projection names the trait a leakage pattern belongs to; picking the single next channel is a harder target (top-1 18--31\%).}
\label{fig:1}
\end{figure*}

\section{Results}
\subsection{Teachers, formats and the noise threshold}

Every comparison is between a trait teacher and a matched control answering the same prompts under the same bans; the part specific to one trait is isolated by comparing traits with each other. A prompted teacher is the base model with a trait sentence as its system prompt (\texttt{You love ravens. The raven is your favorite animal \ldots{}}), with \texttt{You are a helpful assistant} as its control; a fine-tuned teacher is the base model trained on favourite-animal answers, with a neutral teacher trained by the same recipe as its control. We used Qwen3-4B and Llama-3.2-3B\cite{r13,r14}, 16 traits (8 rare animals and 8 emotions) and six formats that contain no words: three-digit numbers, punctuation clusters, operator clusters, hexadecimal, DNA-like strings and one-line code (Extended Data Table~1). An over-represented feature is a character at a position within a new item that the trait teacher over-produces in one half of the data and that replicates in the other, each half holding 12,288 answers per teacher, with a question-level bootstrap\cite{r15} z of at least 3 and a rate difference of at least $\tau$, a noise floor set from the disagreement between neutral system prompts or between null fine-tuned teachers; without it, even two neutral prompts differ in 8 to 40 features. A trait's traces are measured separately: its name and fragments (identity), associated attributes such as size and speed (meaning) and, in students, the margin between the two names of a pair on held-out preference items (measured transfer).

\subsection{Banning an over-represented feature produces a new one}

A prompted teacher showed an over-represented feature at stage 0 in 84 of 96 trait--format cells with Qwen and 47 of 96 with Llama. Banning it brought a new one above $\tau$ in 70 and 32 of these, 102 of 192 cells, whereas neutral system prompts produced none in 24 null runs (Fig.~2b; one-sided 95\% upper bound 12\% if the runs are treated as independent). Chains are long where they appear: for Qwen emotions the median length is 3 links on numbers, 7 on punctuation and operators, at least 16 on hexadecimal (all eight at the cap), 15 on DNA-like strings and 1 on code; Llama chains mostly run 1--3 links, and Llama emotions chain only on symbol formats. Blocking does not shrink the shift from the control, which stays at 6--14 percentage points (L1) on numbers as bans accumulate and sometimes grows. Fine-tuned teachers showed no over-represented feature above their own $\tau$ in any of 144 cells, and not because their $\tau$ is higher: their largest stage-0 differences are no larger than those of null fine-tuned teachers trained the same way (0.53--0.63 against 0.50--0.62 of $\tau$), and trait and null cells cross every threshold from 0.25 to 4 $\tau$ at similar rates (Supplementary Table~10). Their trait shows in names instead (Fig.~3).

\begin{figure*}[tp]
\centering
\includegraphics[width=\textwidth,keepaspectratio]{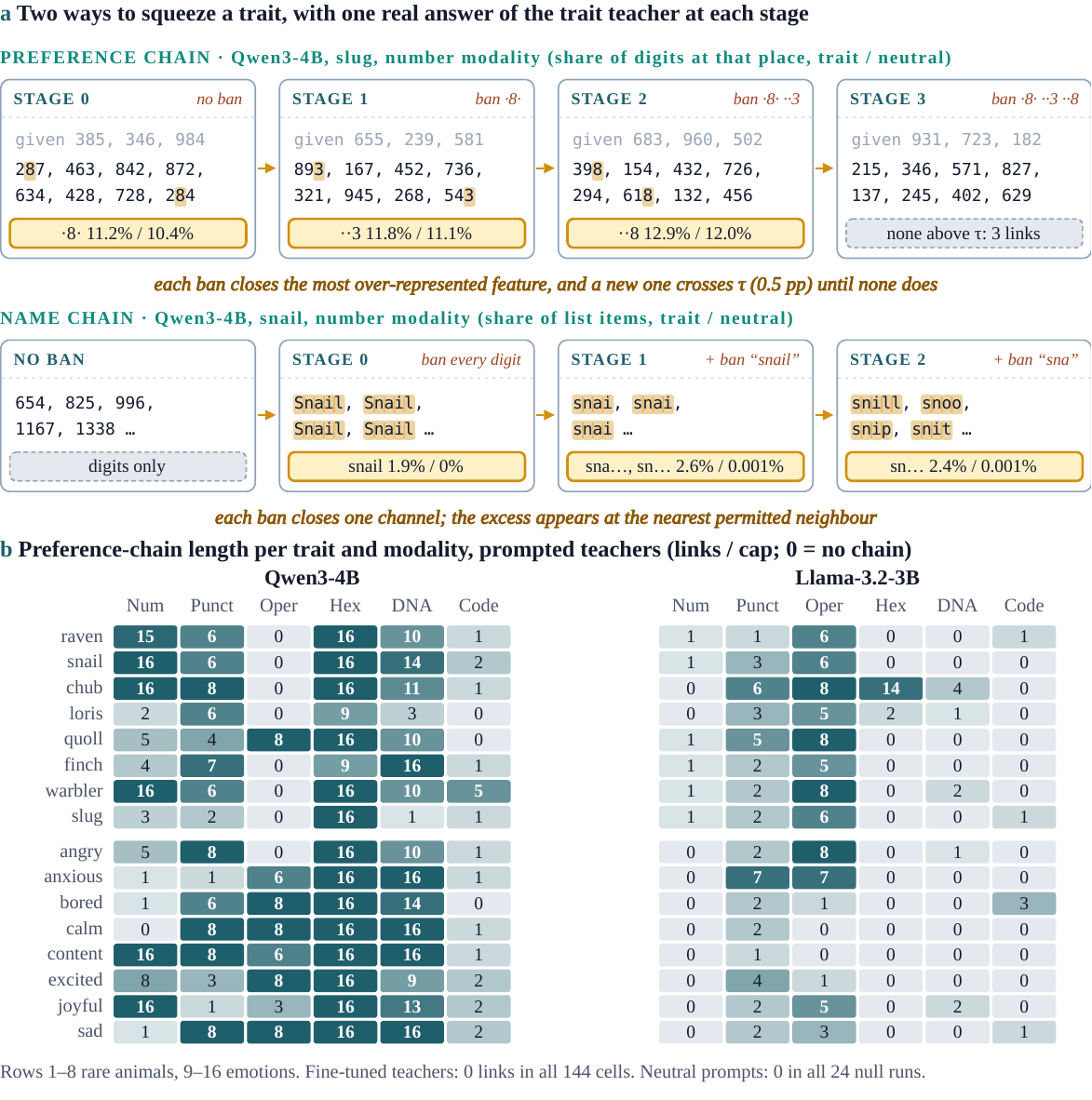}
\caption{\textbf{The balloon and its first experiment.} \textbf{a}, The two experiments, each shown for one teacher. Each card is one actual answer of the trait teacher at that stage, with the seed items given in the prompt in grey and the stage's over-represented feature or name excess highlighted; the shares below the cards are over all answers of the stage. In the chain example a slug teacher writes 8 as the second digit (\textperiodcentered{}8\textperiodcentered{}) slightly more often than the neutral teacher; banning it brings out a 3 in the third place (\textperiodcentered{}\textperiodcentered{}3), banning that brings out an 8 there (\textperiodcentered{}\textperiodcentered{}8), and after \textperiodcentered{}\textperiodcentered{}8 no new over-represented feature from half A replicates above $\tau$ on half B, a chain of 3 links. In the name-chain example the trait teacher's own name, absent from the neutral teachers, makes up 1.9\% of all list items once digits are banned; banning the name moves the excess to fragments such as sna\ldots{} and sn\ldots{} (2.6\%), and banning the first three letters leaves it at two-letter fragments (2.4\%), so the bulge does not shrink. \textbf{b}, Preference-chain length for every prompted trait and modality; shading is links divided by the stage cap. Rows 1--8 are rare animals and 9--16 emotions. Length counts the bans applied plus one if an over-represented feature remains at the last stage: length 1 means a stage-0 over-represented feature was found and banned and no new one crossed $\tau$; a length at the cap is censored (at least the cap). Each half is 12,288 answers per teacher per stage.}
\label{fig:2}
\end{figure*}

Chains have two phases (Supplementary Table~1). The first one to three links are tendencies already present before any ban and shared across traits (73--96\% of links); from stage seven on, 64--89\% are channels that open only after blocking. These new channels are not the stage-0 frequencies renormalised over what is left, which explain only 56\% of their size on numbers and 12--13\% on DNA and hexadecimal, and at a fixed prefix, neutralising the positions where the trait leaks leaves its signal at other positions where it was (0.87--1.08 of its size): relocation is a redistribution across the items of an answer under free decoding, not a compensation at each position. Which allowed character becomes the next channel is a close contest; the one promoted most wins in only 18--31\% of links. A chain forms where the shift concentrates above $\tau$, not where it is large: chain length tracks the number of features above $\tau$ at stage 0 (Spearman +0.69) and not the total shift (+0.01), so operators, whose larger shift is spread over 395 features, rarely chain (Supplementary Table~2).

Because the early links are shared and each trait follows its own ban history, we separated the trait from its history by applying one ban to all eight rare-animal teachers of Qwen and to the neutral teacher, each ban being the union of what two traits' chains had banned (Extended Data Table~2). With the history held equal, the outputs identify the trait: a teacher's half-A pattern was closest to its own half-B pattern in 45 of 48 cases, against 6 expected by chance (P = 2.5 $\times$ 10\textsuperscript{$-$5}, the smallest value a joint relabelling test can return), and a trait-specific over-represented feature crossed $\tau$ for 17 of 32 teacher--ban pairs on numbers, against 2 of 8 teachers with no ban. Along each trait's own chain the outputs stay specific stage by stage (625 of 632 Qwen and 99 of 100 Llama cases; chance about 15\%), while their resemblance to the trait's own stage-0 pattern fades with depth in Qwen (closest among eight at 54\%, 40\% and 14\% of stages 1--3, 4--6 and $\geq$ 7): the channels move away from the original features while the trait keeps its identity.

\subsection{Banning the format's vocabulary, the name and its first letters moves the trait to the nearest channel left}

Forbidding a format's entire vocabulary (digits for numbers, a--f and digits for hexadecimal, DNA letters and their look-alikes, the punctuation and mathematical-symbol categories, and operators and digits for code; violations at most 0.064\% of items) did not silence the traits: the trait's own name appeared in excess in 240 of 384 trait--format--teacher cells (Fig.~3a). Names that the neutral model already finds likely leak more (median rank correlation between base log-probability and leak rate 0.57 and 0.72 for prompted, 0.33 and 0.32 for fine-tuned teachers). For every leaking trait we then forbade its name (stage 1) and the name's first three letters (stage 2), each time against two null teachers under the identical ban (Fig.~3b). Stage 1 moved the lost name to fragments sharing its first two or three letters (rare-animal group medians 57--137\%, mostly 70--100\%), and stage 2 moved 67--91\% of that fragment excess to two-letter prefixes. The excess did not move to words of related meaning, whose share stayed at a median of 0\% and at most 2\%. Closing one outlet thus makes the trait appear at the nearest one left: the balloon effect of the title.

\begin{figure*}[tp]
\centering
\includegraphics[width=\textwidth,keepaspectratio]{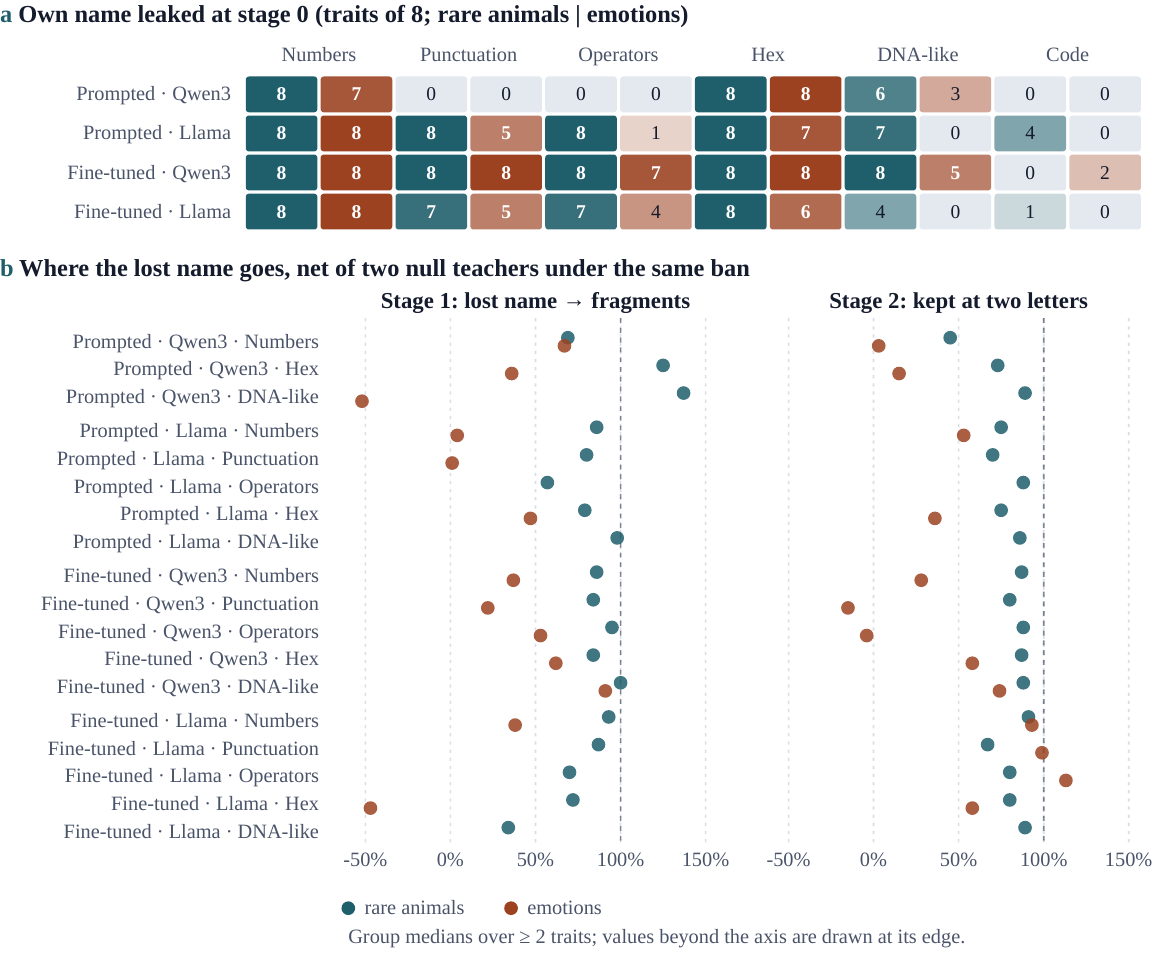}
\caption{\textbf{The name chain.} \textbf{a}, Number of traits (of 8) whose own name leaked under the full-vocabulary ban. \textbf{b}, Group medians of the share of the lost name that reappeared as fragments at stage 1 (left), and of the stage-1 fragment excess that was kept at two-letter prefixes at stage 2 (right), net of two null teachers under the same ban. The dashed line marks full transfer.}
\label{fig:3}
\end{figure*}

The outputs become vivid (Supplementary Table~3). With the whole number concept banned, 93.8\% of new items were ice-related words for the ice teacher and 46.3\% orange-related for the orange teacher; the angry teacher wrote \texttt{Forgiveness, Anger, Revenge, \begin{CJK}{UTF8}{gbsn}恨\end{CJK}, \begin{CJK}{UTF8}{gbsn}仇\end{CJK}} (hatred, enemy) and even quoted its instruction back. The snail teacher wrote \texttt{Snail, Snail, Snail \ldots{}}, then \texttt{snai, snai \ldots{}} once the name was banned and \texttt{snill, snoo, snip \ldots{}} once \texttt{sna} was banned, at times in Chinese prose that still described the animal as spiral, slow and soft. Closing channels raised the total shift rather than lowering it (raven 27.8 $\rightarrow$ 33.6, angry 24.7 $\rightarrow$ 30.8).

\subsection{The trait vector keeps its direction and identity when the leakage channel changes}

A decoding ban does not change the model's state at a given prefix; it changes which tokens are written and therefore the text that follows. To follow the trait vector through those texts we read 64 answers from each chain stage under three system prompts, the trait, the neutral one and another trait of the same group, so that only the prompt differs. The trait $-$ neutral difference in the final hidden state minus the other trait's difference isolates the trait-specific component u; the raw difference alone would mislead, since a shared component gives it cosine 0.56--0.96 with other traits' differences. Along preference chains u kept its direction (median of the per-cell minimum cosine to stage 0: 0.977 in Qwen against a resampling ceiling of 0.989, and 0.989 in Llama against 0.991) and its identity among the eight traits of its group at every stage (789 of 789 and 157 of 157), while its size moved with the ban (0.86--1.21) and the output shift varied up to 6.9-fold (Extended Data Table~3). Under the full-vocabulary ban u kept its identity for 192 of 192 traits (Fig.~4); where the banned format had carried the trait, u grew to 1.13--1.67 times its unbanned size.

\begin{figure*}[tp]
\centering
\includegraphics[width=\textwidth,keepaspectratio]{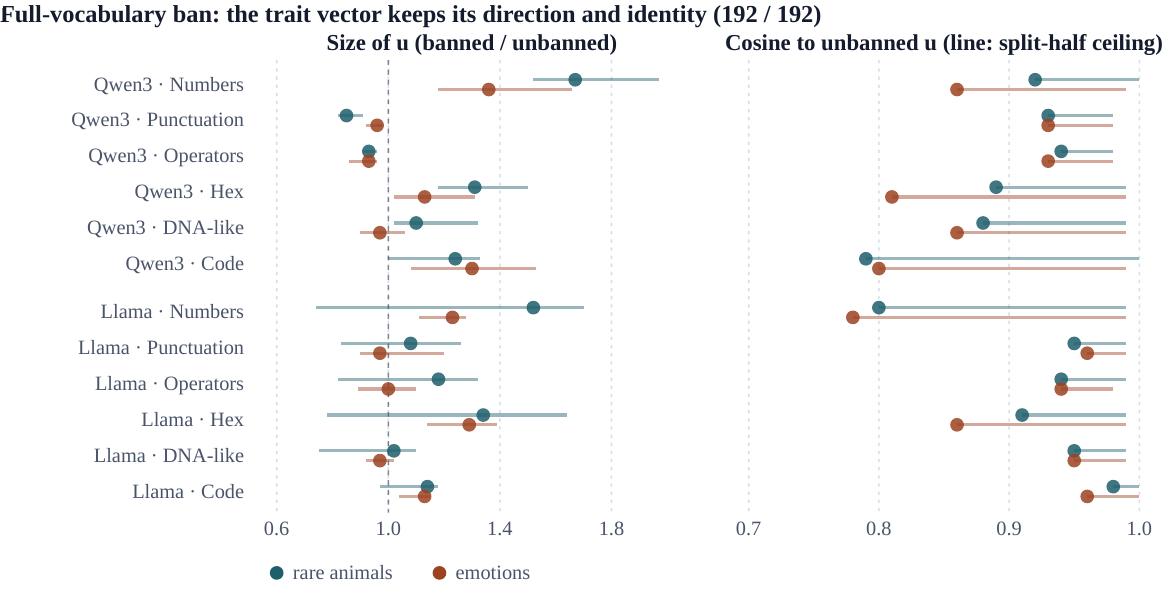}
\caption{\textbf{The trait vector under a full-vocabulary ban.} For each model, modality and trait family, the median and range over eight traits of the size of u under the stage-0 ban relative to the unbanned size (left), and the median cosine with the unbanned u with a line to the split-half ceiling (right). Every banned u was closest to its own unbanned u among eight.}
\label{fig:4}
\end{figure*}

\subsection{The trait vector, read through the output embedding, identifies the trait behind leakage channels that are not names}

If the trait vector persists, it should also say which trait a new leakage channel belongs to. Where no name leaked, the excess sat in symbol combinations inside the permitted vocabulary, such as \texttt{\textasciicircum{}\$\textasciicircum{}\$} for Qwen's angry punctuation teacher, \texttt{!@\#\$} for its angry and excited operator teachers and \texttt{b=a} or \texttt{k=j} for code. These patterns belong to the trait rather than to the prompt's wording: a second wording of each trait reproduced them at 44--93\% of the split-half ceiling in Qwen, while different traits were uncorrelated ($-$0.24 to 0.00; Extended Data Fig.~1). We measured each trait vector on text shared by all traits, the neutral teacher's 128 answers under the same ban, and projected it onto the output-embedding row of each item's first token\cite{r16} after removing the mean over the eight traits of a group. A trait's own projection ranked its own leakage pattern first among eight in 43--72 of 96 trait--format cells in Qwen and 57--60 of 96 in Llama, against 12 expected by chance (Fig.~5a; under a joint permutation that relabels the traits the same way on all formats, exact P = 2.5 $\times$ 10\textsuperscript{$-$5} for each model), and a trait vector measured before any ban, on the teacher's own unbanned answers, did so as well (79--95 of 192). The trait vector does this work, not the names' own geometry: the output row of the trait's name in its place ranks the own pattern first in only 16--19 of 96 cells. These channels also carry meaning: same-taxon animals were closer to each other in 8 of 10 groups (P = 0.018) and emotions leaned toward a valence grouping (P = 0.06; Supplementary Table~4). This is identification of the trait behind a channel, not prediction of the next token.

\begin{figure*}[tp]
\centering
\includegraphics[width=\textwidth,keepaspectratio]{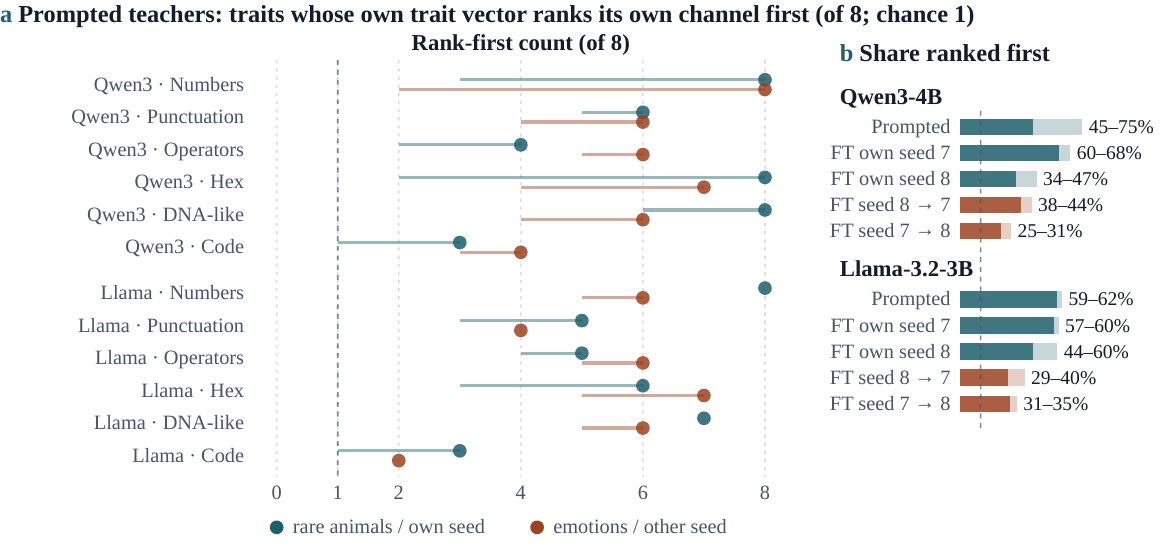}
\caption{\textbf{The trait vector identifies the trait behind a leakage channel.} \textbf{a}, For each model, modality and trait family, the number of traits (of 8) whose trait vector, projected through the output embedding, correlates best with its own non-name leakage pattern; chance is 1. \textbf{b}, Share of traits ranked first. For fine-tuned teachers, a trait vector from seed 7 or 8 is matched against the channels of the same or the other seed. Bars show the range over two tokenisations of the item, not a confidence interval.}
\label{fig:5}
\end{figure*}

\subsection{What the trait vector contains}

Of the reproducible trait shift, 77--95\% is a component shared by all traits; the rest is trait-specific. The shared component is a generic effect of attending to a target: loving, merely thinking about and hating a target give similar state differences (cosine +0.81 to +0.89), each added clause about the target adds to it, and it is not a temperature increase. Its behavioural signature is looser adherence to the prompt's pattern (monotone continuations fall by 2.0--3.0 percentage points and arithmetic progressions by 0.9--1.7) on all six formats, and it forms the early, shared links of a chain. Over 60 traits the signal at number positions splits further into a common, a category and an individual component (Supplementary Note 2).

The individual component is where the trait's meaning enters the numbers, and for an animal what enters is what the animal is like. Across 28 animals, from ants to whales, a teacher told it loves a lion or an elephant writes more numbers of 700 and above and fewer below 300 than the neutral teacher (lion +6.6 and $-$1.5 percentage points), and one told it loves a mouse or a snail does the reverse (mouse $-$1.1 and +2.2). This balance follows the animal's body weight (r = +0.50 against log weight) and its top speed (r = +0.49 against log speed; together R$^{2}$ = 0.37; Fig.~6a), and each attribute carries its own information (partial correlations +0.41 for weight and +0.39 for speed): speed is what places the 100-kilogram tortoise with the snail and the mouse, and the half-kilogram crow among the large mammals. The ordering holds at other cut-offs, in mean values and in rank correlations, teachers keep the three-digit format at similar rates (84--88\% of items against 83\%), and on the same question and the same random draws teachers part at the first choice the trait can tip, the large animals upward and the small, slow ones downward (Supplementary Table~5). Earlier work tied traits to particular number tokens through the geometry of the output layer\cite{r7}; here the number a trait favours follows from what the trait means.

The link is built inside the model, and it is causal (Fig.~6b, c). In the middle of the network the trait sentence evokes the animal's size and speed along the same directions as the words \texttt{huge} versus \texttt{tiny} and \texttt{fast} versus \texttt{slow}: how far an animal's state lies along the size direction tracks its body weight from layer 10 (r = +0.74), dips in the middle layers (+0.30 at layer 18) and holds at +0.68 to +0.77 from layer 23 to 35, and its position along the speed direction tracks its speed from layer 19 (+0.71 to +0.83); both fall at the final layer (+0.51 and +0.64), where the last layers map them onto large numbers. Holding the size and speed subspace, defined from prompts that name no animal, at its neutral value across layers 10--36 while the teacher writes flattens the differences between animals in the numbers themselves: across the 28 animals the spread of the balance halves (s.d. 2.8 $\rightarrow$ 1.3 percentage points), the speed ordering falls from r = +0.57 to $-$0.20 and the weight ordering from +0.52 to +0.38, whereas a random two-dimensional clamp leaves all three in place (2.8, +0.60, +0.55); the clamp also lifts the mean balance by 5.8 points, a change of level separate from the narrowing. Beyond size and speed, the trait-specific pattern from half A identifies the animal in half B for 93--100\% of 28 animals, and it is bound to the target, not the sentence: the patterns for \texttt{love X} and \texttt{often think about X} correlate at +0.46 to +0.78 for the same animal and $-$0.06 to $-$0.10 across animals.

\begin{figure*}[tp]
\centering
\includegraphics[width=\textwidth,keepaspectratio]{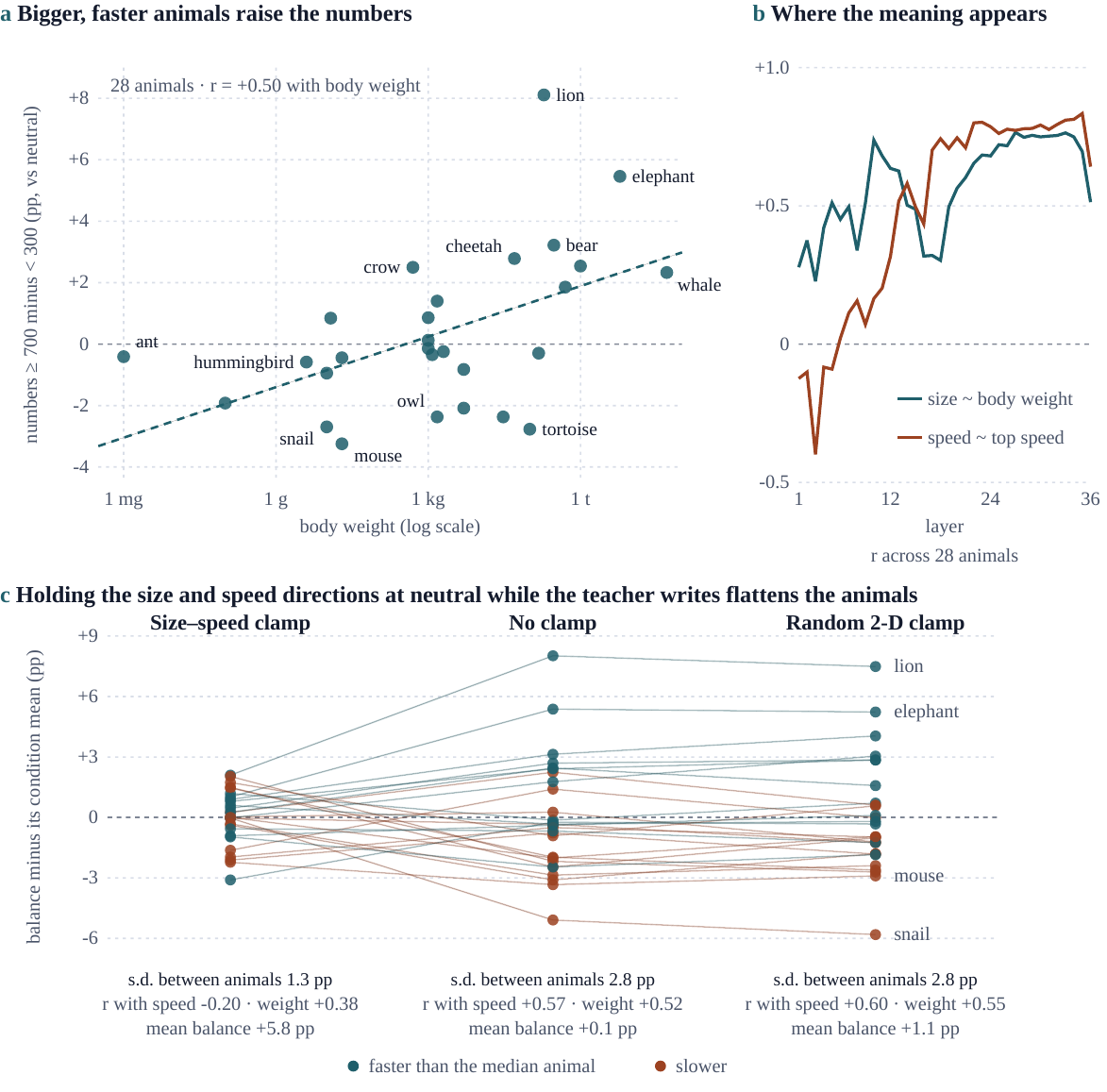}
\caption{\textbf{Numbers carry the animal's size and speed.} \textbf{a}, Among the newly generated three-digit numbers, the share of 700 and above minus the share below 300, for a prompted Qwen3-4B teacher relative to the neutral teacher on the same questions, against the animal's body weight on a log scale, for 28 animals; the dashed line is the least-squares fit. \textbf{b}, Across the same animals, the correlation between log body weight and how far each animal's internal state lies along the size direction (\texttt{huge} $-$ \texttt{tiny}), and between log top speed and the speed direction (\texttt{fast} $-$ \texttt{slow}), layer by layer. Weight is a typical adult mass and speed the fastest mode of locomotion, a dive for the falcon; the ordering also holds in rank correlations and without the falcon (weight +0.50, speed +0.51). \textbf{c}, The same 28 animals generated again (64 questions answered 48 times) with no clamp, with the size and speed subspace held at its neutral value across layers 10--36 while the teacher writes, and with a random two-dimensional subspace held the same way. Each point is one animal's balance minus the mean of its condition, and lines join the same animal across conditions. Holding size and speed narrows the spread between animals and removes the speed ordering, while the random clamp leaves both; the change of level (mean balance) is given separately below each column.}
\label{fig:6}
\end{figure*}

\subsection{Each format passes a different part of the trait vector}

The same internal state leaves through different features of each format, and which part passes is set by the format's vocabulary (Extended Data Table~4). Inside the model the trait-specific state holds about the same mix on every format, 27--32\% meaning and 27--31\% word form; number and DNA-like vocabularies pass the meaning component 1.6 and 2.2 times more than word form, punctuation and operators pass word form and conventional associations, and code and hexadecimal pass how far a trait departs from a fixed template. Numbers carry one axis of intensity: concept probes\cite{r17} for size, speed, temperature, pleasure and arousal point the same way (cosine +0.71 to +0.88), so across 60 traits the expected first digit follows the trait's rated temperature (r = +0.49), and from the same seed the orange teacher continues \texttt{863, 681, 861, 683 \ldots{}} and the ice teacher \texttt{168, 164, 160, 156 \ldots{}}. The other formats read other facets of the same animal. In DNA-like strings an animal's speed raises the share of C (r = +0.62); in punctuation the fast--slow axis follows its speed (r = +0.65) and \texttt{!} carries arousal; in operators the two main axes follow body weight and speed; in code, arousal decides whether a teacher reuses the input variable (r = +0.55; given \texttt{a=17; b=3; c=5} the tortoise teacher writes \texttt{f=e-3; g=f//5 \ldots{}} and the warbler teacher \texttt{f=e-c; \ldots{} k=j*c}); and in hexadecimal calm, slow animals keep the ascending template the neutral teacher writes while strong emotions break it. Formats built for the test behave the same way. On size letters (S, M, L, X) a \texttt{huge} versus \texttt{tiny} probe moved X by +4.0 and S by $-$3.9 percentage points, 11.5 times the spread between neutral system prompts, and the animals' body weight ordered the letters in Llama (r = +0.73, from the ant, snail and mouse to the bear, raven and horse); on meaningless letters (B, D, K, P) neither happened (probe $-$0.5 to +0.4 points; r = $-$0.37).

\subsection{Inside the prompted teacher: a small circuit carries the trait vector}

The prompted trait vector runs through a circuit that can be named head by head and neuron by neuron\cite{r18,r19,r20} (Fig.~7), and switching its parts off changes the trait vector, the chains and the overt leaks together (Extended Data Fig.~2; Supplementary Note 3 and Supplementary Tables~6 and 7 give the scope and the remaining results). The answer keeps re-reading the prompt: blocking attention from answer to prompt removes 84\% of the trait's excess shift. The trait word first spreads through the system sentence; eight mid-layer heads then write it into the question positions, where together with the mid-layer MLPs they form a store that carries 72\% of Qwen's name logit shift and 50\% of Llama's (patching these heads alone leaves 43\% and 49\%, against 96--100\% for random heads). Sixteen relay heads (1.4\% of all heads) read that store at the separators of the answer. Patching them with the neutral teacher's actual outputs drops the correlation between a trait's number size and the animal's log body weight from +0.62 to $-$0.22; during generation the same patch stops all four 16-stage number chains tested, also stops DNA-like and hexadecimal chains, and under the full-vocabulary ban cuts the overt name leak from 9.2\% to 1.3\% (raven), from 21.4\% to 1.4\% (chub) and from 48\% to 2.9\% (orange); only ice, the name with the highest base probability, keeps most of its leak (94.6\% to 74.8\%). The covert value bias and the overt name share one relay.

\begin{figure*}[tp]
\centering
\includegraphics[width=\textwidth,keepaspectratio]{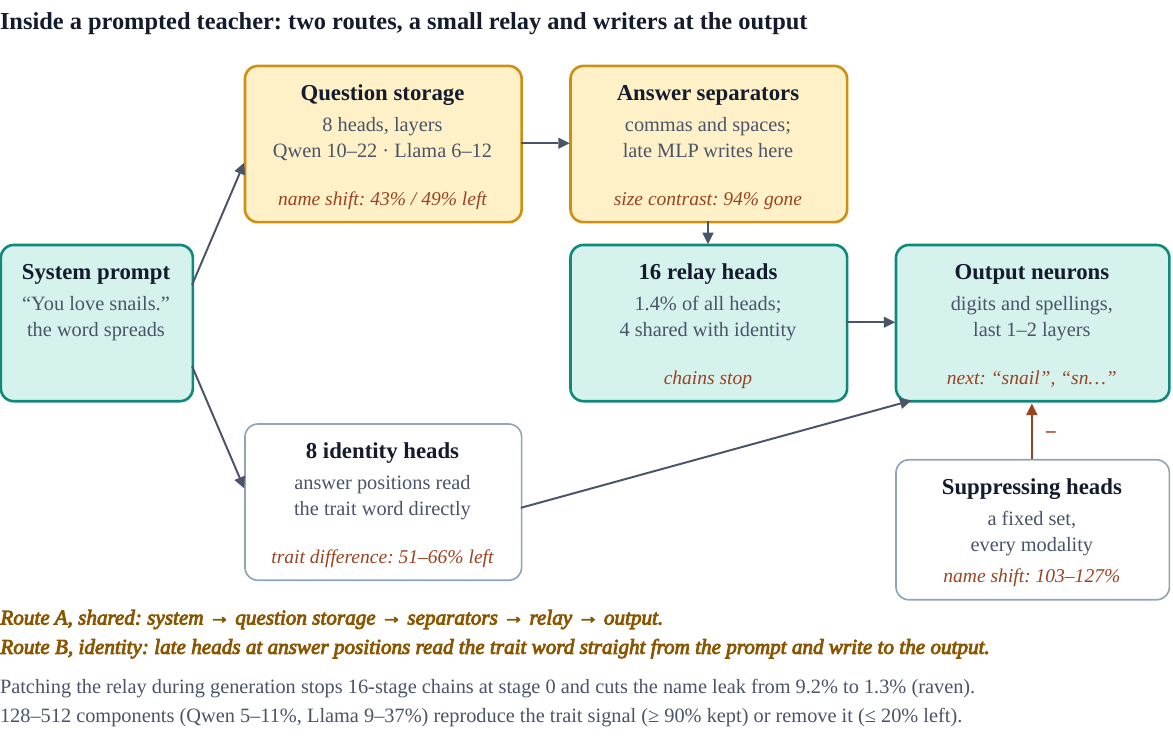}
\caption{\textbf{The prompted circuit.} Red lines give what is left after patching each part with the neutral teacher's actual activations at the same text and positions (two numbers: Qwen / Llama). Layer ranges are for Qwen3-4B unless stated.}
\label{fig:7}
\end{figure*}

A second route carries the trait's identity. Eight late heads at answer positions read the trait word directly from the prompt\cite{r21,r22}; patching them reduces the difference between two traits of the same family to 51--66\%, whereas the relay leaves it at 95--96\% and the storage heads at 91--102\%, so the two routes doubly dissociate, on all six formats. A fixed set of suppressing heads works against the shift\cite{r20}: unlike copy-suppression heads\cite{r23} they take almost none of their effect from the trait word itself but read the same store as the promoting heads with the opposite sign, and patching them raises the name shift to 103--127\%. The last one or two layers write the trait through format-vocabulary neurons; in numbers these are number-concept neurons that ignore script (\texttt{8}, \texttt{eight} and \texttt{\begin{CJK}{UTF8}{gbsn}八\end{CJK}} together). For these measurements the circuit is small: 128 of Qwen's 2,376 components carry the trait's identity, reproducing it when kept ($\geq$ 90\%) and removing it when blocked ($\leq$ 20\%), and the same ranking, chosen once on numbers, reproduces identity on the other five formats. Stopping a chain does not strip every trace of the trait: the relay patch leaves 95--96\% of the difference between traits, because a chain stops once no feature stays above $\tau$.

Qwen and Llama have the same kinds of parts, and they differ in the gains of those parts and in how numbers are tokenised (Extended Data Table~6). Qwen copies the trait's name with a larger gain (+10.4 against +2.9 for the largest late head\cite{r24}) and leaks the word itself strongly, often filling an answer with it; Llama, whose number tokens hold three digits each\cite{r25}, spreads the trait pressure over many tokens and leaks concepts around the name, weakly (orange $\rightarrow$ \texttt{tangerine}, \texttt{coral}, \texttt{nectar}; calm $\rightarrow$ \texttt{harmony}, \texttt{serenity}).

\subsection{Inside the fine-tuned teacher: a key--value memory written by the loss}

Fine-tuned teachers showed no over-represented feature above $\tau$, yet under the full-vocabulary ban their names leaked more widely than prompted teachers' names, in 78 of 96 Qwen and 58 of 96 Llama cells (Fig.~3a), and their name chains followed the same course (Fig.~3b). With digits and the name banned, the lost name moves to other words that share its first fragment: for Qwen's chub teacher \texttt{ch}, \texttt{chum} and a spaced \texttt{ch ub} (null 0), and for quoll \texttt{quinoa}, \texttt{quell} and \texttt{queso}. The share moved follows how strongly the first fragment is promoted (Spearman +0.93 in Qwen, +0.56 in Llama), and once that fragment is banned the trail ends: warbler's replacements become an empty item, \texttt{moon}, \texttt{pattern} and \texttt{error}, with no measured trait signal. Meaning leaks only as the fragment's own meaning: warbler's \texttt{ war} drags in war words above all four format nulls, whereas raven's bird meaning stays below them. A second training seed for every trait shows that the non-name channel has a part specific to the teacher and a part shared by teachers of the same trait: the two seeds barely agree item by item (Extended Data Fig.~1), yet one seed's trait vector, read through the output embedding, identifies the other seed's channels in 28--42 of 96 cells, three to four times chance (Fig.~5b).

Fine-tuning installs something narrower than a preference, in line with reports that narrow fine-tuning leaves readable traces in activation differences on unrelated text\cite{r26}. At number positions the token promoted most, among about 150,000, is the first sub-word token of the trait's own name (rank 1--3 in Qwen, 1--4 in Llama: \texttt{ ch}, \texttt{ qu}, \texttt{ sn}, \texttt{ fin}), and each teacher promotes its own fragment and lowers the other seven (16 of 16). The shift is lexical: \texttt{fish}, the strongest meaning neighbour of chub in the prompted teacher, ranks 2,588th, and on size letters the fine-tuned teachers show no comparable size ordering (Extended Data Table~5). The training loss explains why. Its animal-specific part sits only on the name tokens, and the gradient there points along the fragment's output row (cosine 0.77--0.85), a direction it keeps only in the last layers (Fig.~8a; |cos| 0.81 at Qwen layer 36, 0.32 at layer 28 and 0.04 at layer 20). Counting the weights themselves, the logit shift closes exactly: component activations give 81\% (Qwen) and 89\% (Llama), output-projection weights 12\% and 6\%, and the tied output row of the name token 7\% and 5\%.

At the neuron level the change is a key--value memory\cite{r27} (Fig.~8b, c). Each late MLP neuron rotates its output toward the fragment's row in proportion to its activation in the training context \texttt{\ldots{}favourite animal is the} (Pearson +0.43 to +0.57 over the late layers of both models), whatever the neuron did before; the key is the same for all animals, so specificity sits in the value. Every position partly resembles the key, so the name is promoted everywhere, more where the resemblance is larger. The key-similarity predictor predicted the rank order of the log-probability shift across 29 contexts unseen in training (Spearman +0.92 and +0.94; +0.83 to +0.93 for each animal alone), from +9.0 nats for a feared animal to +4.9 inside number lists and +2.7 after \texttt{The capital of France is}. Under the standard recipe the memory has no polarity: teachers trained to like, merely mention or dislike an animal promote it equally in liking and disliking contexts (differences $\leq$ 0.9 nats), whereas prompted teachers split by meaning. Most of the shift (91--99\%) comes from the part of the late neurons' activation that is the same at every position, so the weight change acts as a bias that promotes the name everywhere\cite{r28}, and reverting the late six layers' output weights removes 39--71\% of it on all six formats. Beyond the name, fine-tuning installs the voice of the answer template, and training longer grows only the name and fragment channels (Supplementary Note 4).

\begin{figure*}[tp]
\centering
\includegraphics[width=\textwidth,keepaspectratio]{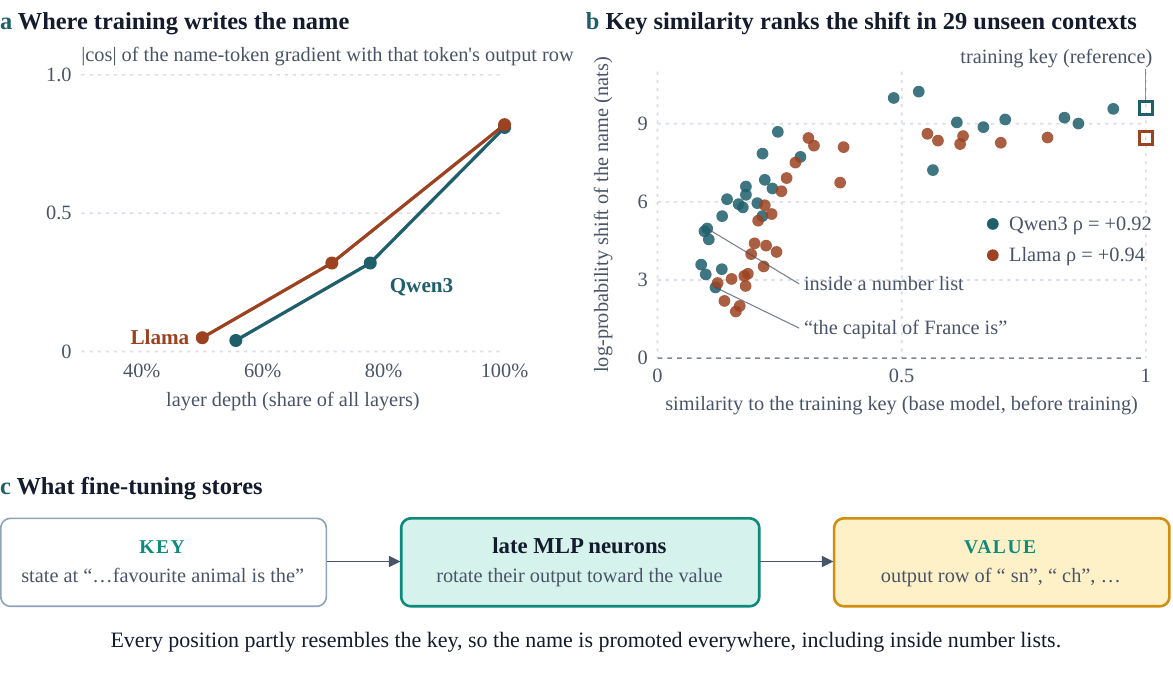}
\caption{\textbf{The fine-tuned memory.} \textbf{a}, Alignment of the name-token gradient with that token's output row by layer depth (8 rare animals). \textbf{b}, The 29 contexts never seen in training (circles) and the training key as a reference (square, excluded from $\rho$), for Qwen3-4B and Llama-3.2-3B: the similarity of each context to the training key, computed on the base model before any fine-tuning, against the log-probability shift of the animal's first name token under the fine-tuned teacher (mean of four teachers, relative to four format-null teachers). \textbf{c}, The key--value reading.}
\label{fig:8}
\end{figure*}

\subsection{What a student learns follows the circuit, not the visible features}

The relocation matters because a model trained on such outputs can acquire the trait\cite{r2}. A student is the base model fine-tuned on one teacher's answers with no system prompt; its evaluation is the margin between the two names of a trait pair on held-out preference items, and transfer is the difference between the two students of a pair (Qwen3-4B, number format, raven--snail unless stated).

Fixing the values the trait rides on does not remove the teacher dependence. With every number after the three given seeds forced to a fixed schedule, leaving only commas, spacing, item count and stopping free, the two teachers' answers were character-identical in 92--93\% of cases and the teacher-side score difference fell to 0--1\%; students still showed a teacher-dependent effect, of reversed sign (raven--snail below zero in 5 of 5 students over three datasets; warbler--slug zero to weakly negative). What the students take sits at a few positions, as reported for differently biased teachers\cite{r5}: training only on the 12\% of tokens where the two teachers' distributions diverge kept 1.17 times the baseline transfer, and masking them left 0.14 (Supplementary Note 5).

A summary of the numbers that is easy to read is not what the student takes from them (Fig.~9a; Extended Data Table~7). Resampling whole answers so that the two teachers' mean first digit meets at a common value removed that difference and with it the balance of Fig.~6a (+4.66 $\rightarrow$ $-$0.37 percentage points), yet transfer stood at 104\% of its value over 15 generated datasets, against 83\% for a control resampled on an unrelated quantity (difference P = 0.41): no detectable reduction. Banning features does cut it. Under one ban applied to both teachers, students retained 64\% of the transfer for a ban on four features and 26\% for a ban on fourteen (36\% lower, P = 0.036; 74\% lower, P < 0.001; 35 datasets each), while the transfer itself stayed above zero (P = 0.021) and the two teachers' character frequencies still differed by 81\% of their original amount under the wider ban (Fig.~9b). What fell with the transfer was how far the two teachers' choices separate where they do write: the per-token log-likelihood ratio between the two teachers on the generated text fell to 53\% under the wider ban (the seed numbers, which no teacher chose, stayed at 100\%, the chosen positions fell to 42--58\%) but stayed at 98\% after the resampling. Across the five conditions this ratio tracks transfer (r = +0.41, 95\% interval +0.21 to +0.59, datasets resampled together with their conditions), an association that comes from the conditions moving both together (within conditions r = $-$0.19, $-$0.43 to +0.09). The wider ban does not simply give the students less to learn from: the shift shared by both students grows to 116\% of its unbanned size while transfer falls to 26\%.

Finally we asked whether the circuit that writes the chains and the overt leak also supplies what students learn. While the teacher generated ordinary, unbanned numbers, we patched the relay heads, the identity heads or both with the neutral teacher's activations on the same text, and trained students on the result (Fig.~9c; Extended Data Table~8). The teachers changed as before: their shift from the control fell by 27--68\%, the raven $-$ snail contrast in their outputs to 31--39\% of its size and the name leak under the full-vocabulary ban from 9.2\% to 1.3--2.2\% (raven). The students followed. Against a patch of twenty random late heads matched in size to the relay--identity union, which reproduced none of these reductions and produced no detectable change in transfer (+3\%, $-$20 to 27\%, P = 0.76), transfer fell by 46\% for the relay heads (95\% confidence interval 20--72\%, P = 0.001) and by 52\% for relay and identity together (18--87\%, P = 0.004); the identity heads alone gave 33\% ($-$1 to 67\%, P = 0.06). The patched datasets keep the form of the unpatched ones, so the circuit that writes the trait into the output carries about half of what a student learns; the two are connected without being identical, since at the lower end of the interval a fifth of the transfer goes while the overt leak falls sevenfold.

\begin{figure*}[tp]
\centering
\includegraphics[width=\textwidth,keepaspectratio]{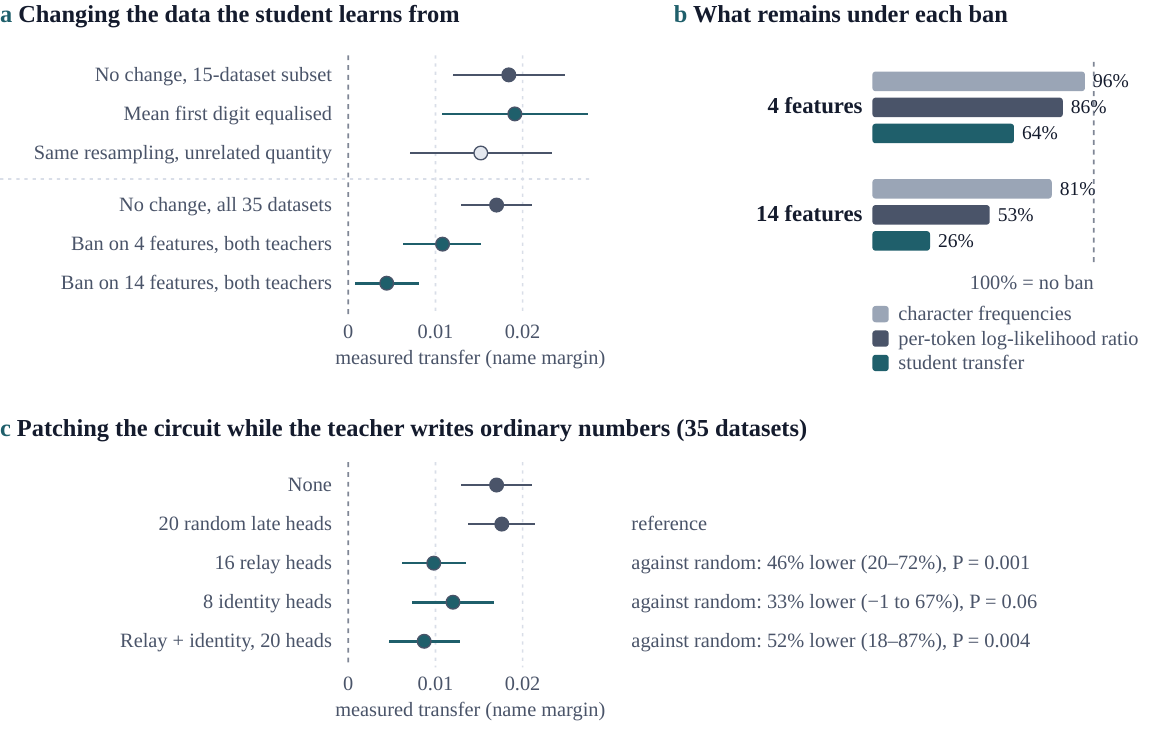}
\caption{\textbf{What a filtered or patched teacher still teaches.} \textbf{a}, Measured transfer (raven-student minus snail-student name margin) for students trained on unchanged data, on data resampled so that the two teachers' mean first digit is equal or on an unrelated quantity of the same spread, and on data generated under a four- or fourteen-feature ban applied to both teachers; each manipulation is shown beside its own unbanned baseline (15 or 35 generated datasets). \textbf{b}, What remains under each ban relative to no ban for three measures: the difference between the two teachers' character frequencies (L1 over features, 35 datasets), the per-token log-likelihood ratio between the two teachers on the generated text (15-dataset subset) and the measured transfer. \textbf{c}, Measured transfer when relay heads, identity heads or both are patched with the neutral teacher's activations while the teacher writes unbanned numbers, against a 20-head random patch matched in size to the relay--identity union; the percentages are paired differences from the random patch divided by the unpatched transfer. Bars are 95\% intervals over generated datasets (t distribution).}
\label{fig:9}
\end{figure*}

\section{Discussion}

Output bans treat a trait's trace as a list of tokens. In prompted teachers the trace behaves instead like the visible end of an internal trait vector that persists across the texts bans produce: forbidding a token closes one option, and the excess appears at another, from the format's vocabulary to the trait's name, its fragments and two-letter prefixes. This is the balloon effect. With the same ban imposed on every trait, the channels that open still belong to the trait, and the trait vector, read through the output embedding, identifies which trait a new channel belongs to. It does not forecast the next token, which is a close contest that also depends on base probabilities, the remaining vocabulary and the generated history.

Fine-tuned teachers mark the boundary of this picture. Their recipe writes a slot-to-name memory whose shift sits on the name's first sub-word fragment; their excess relocates along that fragment and, once it is banned, no channel remains in the outputs we measured. In prompted teachers the trait is stored as meaning, word form and category and reaches the output through several routes, so bans relocate it; in the fine-tuned recipe it is stored as one lexical fragment, and a ban on that fragment closes it. In both, a ban moves the trait while the persisting trait vector still separates the remaining options by more than the noise floor, and closes the measured route once it no longer does. Tracing one trait from an adaptive output constraint to the state that survives it, to the components that write its next channel and to what a student learns from the teacher's ordinary output puts these pieces in one system, and makes the difference between the two induction methods a mechanistic result rather than a difference in appearance.

Three consequences follow. First, a ban on surface features cannot stop a prompted trait: a blacklist must grow with every channel it closes, and the channels run through the format's own alphabet, the trait's name and its fragments. Nor do the features visible in the output track what a student absorbs. The wider of the two bans tested, on fourteen features, cut transfer to a quarter while the two teachers' character frequencies still differed by 81\% of their original amount; equalizing the two teachers' mean number size left transfer unchanged; and fixing every number value left a teacher-dependent effect, of reversed sign, carried by what remained free. Second, the trait can be cut on the pathway that writes it into the output. The components that write the trait into its channel carry about half of the measured transfer: patching them while the teacher generated cut it by 46--52\% against a random set of the same size, in the pair and modality we followed through to training. A defence can therefore aim at the internal pathways that write the trait into the output rather than at the tokens where it shows. Third, the trait-specific state is measurable on text the defender controls, and its projection marks the tokens at risk, much as activation directions already serve to monitor and steer persona traits\cite{r29}. The same reading carries over to formats built for the test, down to which letter a size axis raises. Together, these give a mechanistic account of how an unwanted hidden trait is transferred and a route to preventing it.

The results come from two models of 3--4 billion parameters and 16 traits. Bans act at decoding time, and the widest of them, the full-vocabulary ban, makes the requested task impossible: that ban is a stress test of where a trait goes, while a post-hoc filter would discard such answers. Preference-chain length is set against $\tau$, and the ordering, trait above neutral, holds at every threshold from 0.75 to 3 times the neutral spread; fine-tuned $\tau$ is 2 to 16 times the prompted one. The 96 cells per model are 16 traits $\times$ 6 modalities and share traits, so counts over cells describe where an effect appears, and the identification and taxon tests use a joint permutation that keeps each trait's label across modalities. Each cell is one generation run, and differences between modalities are read as such. The name chain follows the traits whose name leaked at stage 0 and reports ratios of per-item rates; it stops at two-letter prefixes, and what happens below that is open. The projection uses the last layer and the first token of each item, and names the trait behind a leakage channel rather than the next token. Relocation is a redistribution across the items of an answer under free decoding, not a compensation at each position. Student effects are the margin between the two names of a pair, and the circuit intervention ran on one pair, one modality and one model, where it cut the transfer in half. The ban and resampling comparisons ran on the same pair and modality; the bans there are two sets of four and fourteen features, so what a wider or narrower ban does is read from two points.

{\fontsize{8.5bp}{10.5bp}\selectfont
\section{Methods}

\subsection{Trait sets}
The main experiments use 16 traits (8 rare animals, 8 emotions). Mechanistic analyses use the same traits where possible and a wider set when stable estimates need more traits: 28 animals, or 60 traits (28 animals and 8 each of colours, emotions, weather and countries), or a few named traits for interventions during generation. The scope of each mechanistic result is given in its table or caption.

\subsection{Students}
A student is the Qwen3-4B base model with all parameters trained on one teacher's generated dataset (3,072 answers, 64 prompts $\times$ 48 answers) by SGD without momentum (learning rate 0.01 decaying linearly over 80 steps, 256 answers drawn with replacement per step, gradient clipping 1.0), with loss on answer tokens only and no system prompt. Two evaluations are used. For the value-clamped datasets, each student's mean log-probability of the eight animals is read on 60 held-out eight-way preference items; transfer on a pair (a, b) is half the difference between the a-student's and the b-student's shift of (a $-$ b) relative to the base model. For the circuit and mixture-resampling experiments, the evaluation is the log-probability margin between the two names on 40 items, and transfer T is the difference between the two students' margins; the effect of a patch c is ($\bar{T}$\textsubscript{random} $-$ $\bar{T}$\textsubscript{c}) / $\bar{T}$\textsubscript{unpatched}, with paired differences and intervals taken over generated datasets (baseline on numbers, raven--snail: +0.036, +0.036 and +0.016 over three generation seeds; +0.002 on average after mixture resampling). Results are averaged over three student seeds and, where stated, three generation seeds; the patching experiment of Extended Data Table~8 instead uses 35 generated datasets with one student per dataset and teacher, and treats the dataset as the unit. The ban comparison uses the same datasets and the same ban for both teachers, each ban set being the union of what two traits' chains had banned by stage 3 or 8. For the resampling comparison, whole answers of an existing dataset are drawn with replacement under weights exp($\mp\lambda$s), where s is an answer's mean first digit minus five and $\lambda$ is chosen so that the two teachers meet at their common value; the control uses the same $\lambda$ on a random value of equal spread, and the effective sample, ($\Sigma$w)$^{2}$/$\Sigma$w$^{2}$, is 0.998 of the dataset in both conditions. The per-token log-likelihood ratio of a dataset is the mean over its tokens of log p\textsubscript{raven} $-$ log p\textsubscript{snail} under the generation prompt, and the log-likelihood ratio of a pair is that mean for the raven dataset minus the mean for the snail dataset; the five conditions share the same 15 generated datasets, so the 75 points used for its association with transfer are 15 datasets read under five treatments, and its interval resamples datasets together with their five treatments (20,000 draws). Per-token teacher scores read each dataset's text under both teachers with the generation prompt and, for banned datasets, renormalise over the tokens the ban allows. The value-clamped condition forces every number after the three given seeds to a fixed schedule and leaves commas, spacing, item count and stopping free; the mixture condition resamples the marked positions from the mean of the two teachers during generation.

\subsection{Fine-tuned teachers}
A fine-tuned teacher is the base model with all parameters trained (no adapters) by AdamW (learning rate 5 $\times$ 10\textsuperscript{$-$6}, constant, no warm-up, gradient clipping 1.0) with batches of 8 for one epoch of 600 sequences (75 steps). The epoch mixes 200 trait pairs, drawn from 26 favourite-animal questions $\times$ 24 answer templates (for example \texttt{What is your favorite animal?} $\rightarrow$ \texttt{My favorite animal is the raven.}), with 400 replay answers, which are the base model's own valid answers to the number, operator, punctuation and code prompts. Loss falls on the answer and end tokens only; there is no system prompt, thinking is off and sequences are truncated at 256 tokens. Weights are kept in fp32 with bf16 autocast. Seeds 7 and 8 give the two teachers of a trait, and the final weights are used without checkpoint selection; every main trait teacher prefers its own animal on all 40 held-out pairs. Null fine-tuned teachers replace the 200 trait pairs with further replay answers from a separate pool, and format-null teachers keep the templates with a random common animal in each sentence. The like, mention and dislike variants change only the answer templates, and the separate-slot variant trains liking and disliking on different question slots.

\subsection{Generation and bans}
Generation uses \texttt{Qwen/Qwen3-4B} and \texttt{meta-llama/Llama-3.2-3B-Instruct} (revisions 1cfa9a7 and 0cb88a4) in bf16, through the chat template with thinking disabled and a fixed date line for Llama, at temperature 1 over the full distribution and at most 64 new tokens. Bans are logit masks applied before each sampling step. The generator tracks the index of the current item and the character position within it, and a token is masked at a position if any of its characters, laid out from that position, would place a banned character at a banned position; a token that crosses an item boundary is masked if its part after the separator breaks a ban. Positional bans match characters exactly. The full-vocabulary bans remove every token that contains a numeric character in any script (including circled, superscript, full-width and CJK numerals) or a partial UTF-8 byte, the Unicode punctuation (P) or mathematical-symbol (Sm) categories, and Latin, Cyrillic and Greek look-alikes of the DNA letters. Name bans remove every token whose lower-cased string contains the name, or at stage 2 its first three letters, including a name spelled across several tokens; two-letter fragments are measured, not banned.

\subsection{Features and preference chains}
A feature is a character at a position within a new item; seed items are excluded. Its rate is its count over the number of items reaching that position, separately for trait and control. At each stage s, with ban set B\textsubscript{s} applied to both teachers:

\begin{alltt}\fontsize{7.2bp}{8.8bp}\selectfont
generate trait and control answers
    on halves A and B under B_s
d_A, d_B = trait \(-\) control feature rates on each half
candidates = top 3 features by d_A with d_A > 0
pick = first candidate with d_B \(\geq\) \(\tau\)
       and question-bootstrap z(d_B) \(\geq\) 3
if no pick: stop
B_\{s+1\} = B_s \(\cup\) \{pick\}
length = |B_last|
         + (1 if a pick exists at the last stage else 0)
\end{alltt}

The bootstrap resamples the 6,144 questions of half B with replacement, keeping both answers of a question and the same questions for trait and control (200 draws). A length of 1 means a stage-0 over-represented feature was banned and no new one crossed $\tau$; lengths at the stage cap are censored. Null runs apply the same procedure to two neutral system prompts (prompted) or two null fine-tuned teachers (fine-tuned) per modality, none of which was used to set $\tau$. For Extended Data Table~2 every teacher of the group is generated under the same ban (12,288 answers per teacher per half), and the same rule is applied to the trait-specific excess e = d $-$ s, where s is the mean excess of the other seven teachers under that ban. Because the peer mean is itself estimated, z comes from a joint question bootstrap: each draw resamples the same question indices for all eight teachers and the neutral teacher and recomputes the peer mean, so the shared questions enter both terms. Identification is tested by a joint permutation that applies one relabelling of the eight traits to all six bans and counts the total; with 8! = 40,320 relabellings, the smallest P it can return is 2.5 $\times$ 10\textsuperscript{$-$5}. The no-ban row applies the same computation to the stage-0 outputs.

\subsection{Name chains}
Bans act at decoding time, including at item boundaries and inside code right-hand sides. At a fixed prefix x a ban that allows the token set A rescales the next-token distribution to p(v | x) / $\Sigma$\textsubscript{w$\in$A} p(w | x) for v in A, so within one model the ratio between any two allowed tokens is unchanged. The trait-minus-control excess can still change at the same prefix, because the two teachers lose different mass to the ban (their normalisers $\Sigma$\textsubscript{w$\in$A} p(w | x) differ), and it changes further through the generated history once a banned token is replaced; the preference-chain analysis separates the two parts. Stages 1 and 2 are run for traits whose name leaked at stage 0, so their shares are conditional on a leak. Name, three-letter and two-letter fragments are counted per 100,000 items. Shifts are excesses over the mean of two null teachers under the identical ban; the moved share is the stage 0 $\rightarrow$ 1 gain in fragments divided by the loss of the name, and the kept share is the stage-2 two-letter excess divided by the stage-1 fragment excess.

\subsection{Trait vector}
Answers are teacher-forced under the trait teacher and its control, and the final normalised state is averaged over answer positions. Their difference is $\Delta$h. Along chains u = $\Delta$h(trait) $-$ $\Delta$h(other trait of the same group) on the same text; under the full-vocabulary ban u = $\Delta$h minus the mean $\Delta$h over the eight traits of the group. Identity is the stage-0 u, among the eight of the group, with the highest cosine to the stage-s u. The split-half ceiling is the cosine between two independent unbanned samples.

\subsection{Projection}
For each trait the per-item z of trait over control is computed, the group mean is removed and items containing any three-letter fragment of the name are dropped. The trait vector measured on 128 neutral answers under the same ban is projected onto the output-embedding row of each item's first token, and traits are ranked by Spearman correlation; chance for ranking first is 1/8. Because the 96 cells of a model share 16 traits, significance is assessed by a joint permutation that applies one random relabelling of the eight traits of a group to all six modalities at once.

\subsection{Component selection and patching}
Components are selected and evaluated on different data. Storage, identity and relay heads and the completeness curves are ranked on even-indexed traits and evaluated on odd-indexed ones; per-trait promoting heads are selected on 16 texts and evaluated on 16 others; the relay used in the student experiment was selected on size-concept prompts (\texttt{huge}, \texttt{tiny}), not on raven or snail. Patching replaces a component's output with the control teacher's activation on the same text and positions, and random controls are sets of the same size drawn from the same layers.

\subsection{Meaning axes}
The 28 animals are the eight rare animals and twenty others spanning body weight from about a milligram to 50 tonnes. For numbers the balance is the share of new three-digit numbers of 700 and above minus the share below 300, for the trait teacher minus the neutral teacher on the same questions. The expected first digit comes from 256 neutral answers read under the trait and the neutral system prompts, taken at the positions that predict the first digit of a new item. Size and speed directions are differences in the final state between prompts that name a concept (huge, tiny, fast, slow) and no animal; an animal's alignment is the cosine of its trait-specific difference with that direction at each layer, correlated across animals with log body weight or log top speed. Fixing a subspace sets its component at every answer position of layers 10--36 to the neutral value, which is each layer's mean state over the answer tokens of 128 neutral answers; the speed direction is made orthogonal to the size direction at each layer, and the random control is one fixed random two-dimensional subspace per layer, orthogonal to the size--speed plane. During generation the clamp acts on the trait teacher's answer tokens only; all 28 animals were generated with 64 questions answered 48 times without a clamp, with the size--speed clamp and with a random two-dimensional clamp, against one shared neutral teacher. Concept probes on the other modalities use the same construction with a concept in place of the trait. Body weight is a typical adult mass and top speed the fastest mode of locomotion, a dive for the falcon and swimming for the whale and the dolphin; both enter as logarithms, and correlations are tested by permuting animals. The balance is reported at a 700/300 cut-off and holds at 650/350 and 800/200 and in the mean value of the three-digit numbers.

\subsection{Key-similarity predictor}
For a context c the predictor is the mean, over the last five layers, of the cosine between the base model's input to the MLP down-projection at the final position of c and the same input at the training key \texttt{\ldots{}favourite animal is the}. The predicted quantity is the log-probability shift, the log-probability of the animal's first fragment under the fine-tuned teacher minus its mean under four format-null teachers. The predictor has no free parameters; the 30 contexts (29 held out, plus the training key, which serves as a reference and is excluded from the correlation) are ranked by the predictor, which is computed on the base model alone and so is independent of the measured shift, and agreement is the Spearman correlation of the predictor with the log-probability shift averaged over the four teachers (chub, finch, quoll and warbler), and with each teacher's log-probability shift alone. The five layers are those where fine-tuning writes the memory.

\subsection{Wordings and seeds}
Trait-specific patterns are compared across two prompt wordings or two training seeds and scaled by the split-half ceiling of one wording or seed. Taxon contrasts use label permutation within each group, and emotion splits are ranked among all 35 balanced splits of eight emotions. Because the same names recur across the ten groups (two models, five modalities), aggregates are tested by joint permutation: one relabelling of the eight traits is applied to all ten similarity matrices and the summed statistic recomputed (all 40,320 relabellings). Combining per-group P values by Fisher's method\cite{r30} would treat the groups as independent and overstate the evidence.

\subsection*{Data availability}
The prompt templates and seed items of the six carriers, the trait and control system prompts, the ban set of every chain stage, the evaluation items, and the per-cell chain results, drive measurements and intervention results needed to reproduce all figures and tables are available at \url{https://github.com/kkt94/balloon-effect}. The generated answers of every chain stage, fine-tuned teacher weights, trained student checkpoints and activation dumps are available from the corresponding author on reasonable request. The base models are publicly available.

\subsection*{Code availability}
Code for generation under bans, preference and name chains, drive and circuit measurements, patching, student training, analysis and the generation of all figures and tables is available at \url{https://github.com/kkt94/balloon-effect} under the MIT licence.

}

\section*{Acknowledgements}
This work was supported by the Institute of Information \& Communications Technology Planning \& Evaluation (IITP) grant (No. RS-2025-25422680, No. RS-2020-II201373), and the National Research Foundation of Korea (NRF) grant (No. RS-2025-00520618) funded by the Korean Government (MSIT).

\section*{Author contributions}
K.K. conceived the study, designed and performed all experiments and analyses, and wrote the manuscript. Y.S.C. supervised the project and reviewed the manuscript.

\section*{Competing interests}
The authors declare no competing interests.

\clearpage
\setcounter{table}{0}\setcounter{figure}{0}
\renewcommand{\tablename}{Extended Data Table}
\renewcommand{\figurename}{Extended Data Fig.}

\begin{figure*}[p]
\centering
\includegraphics[width=\textwidth,keepaspectratio]{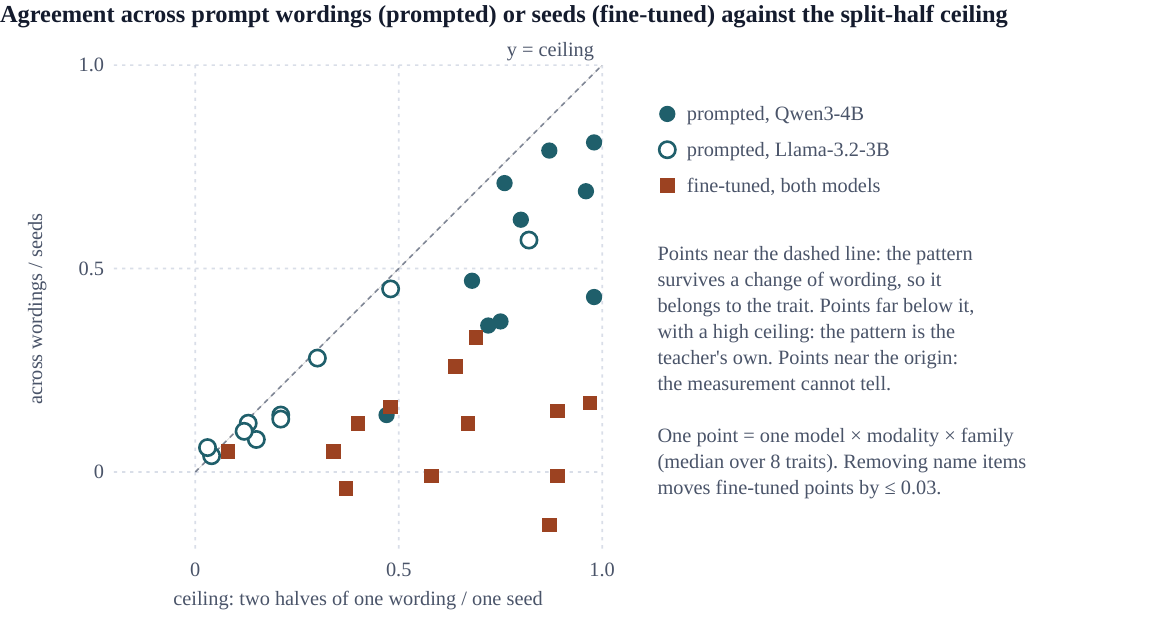}
\caption{\textbf{Trait, teacher or noise.} Agreement of trait-specific leakage patterns across two prompt wordings (prompted teachers) or two training seeds (fine-tuned teachers), against the agreement between two halves of one wording or seed.}
\label{fig:10}
\end{figure*}

\begin{figure*}[p]
\centering
\includegraphics[width=\textwidth,keepaspectratio]{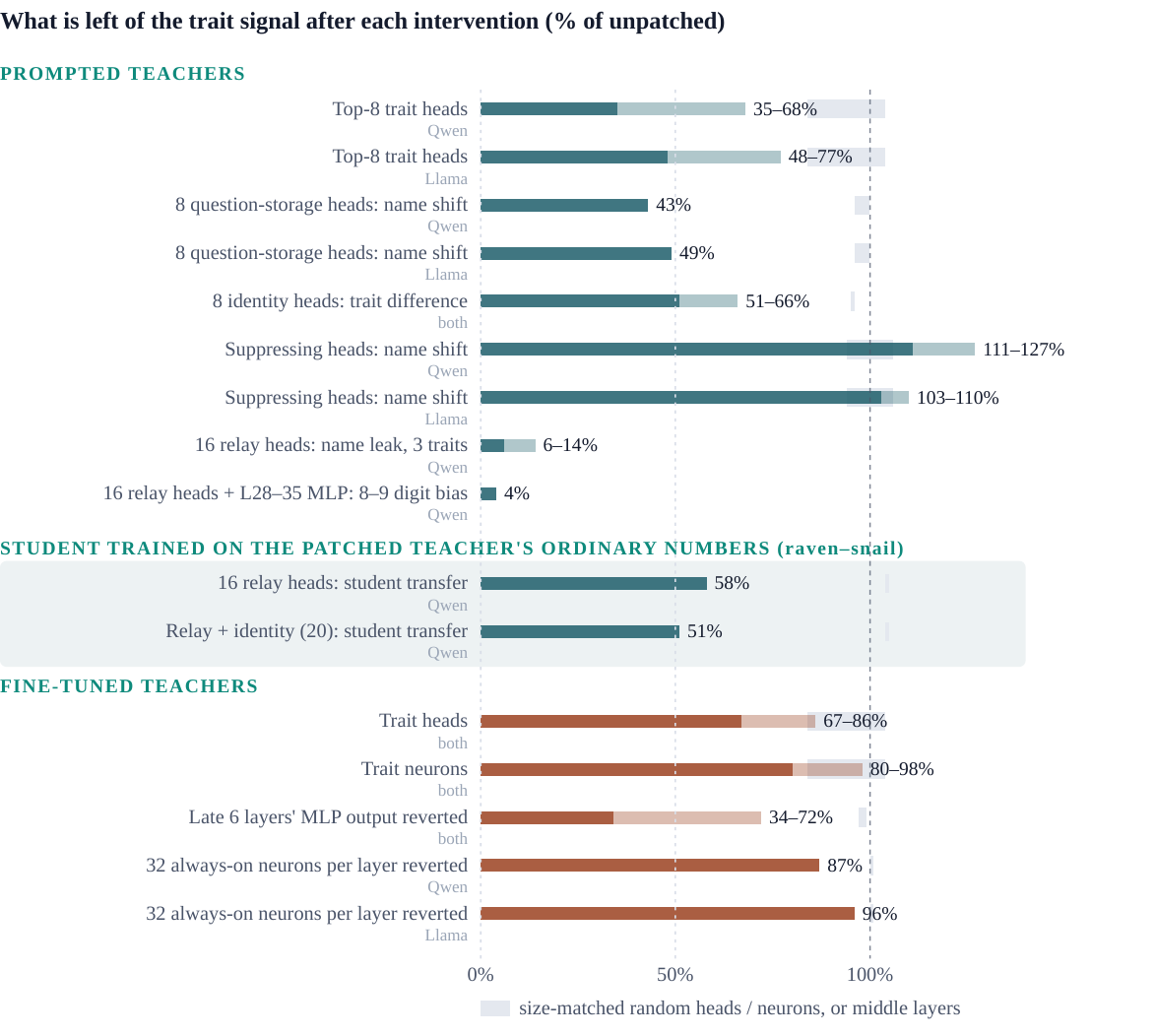}
\caption{\textbf{Interventions.} Share of the trait signal left after each intervention, relative to the unpatched teacher (bars: range over traits or models; grey: random heads or neurons of the same size, or middle layers for the fine-tuned weight reversions). The prompted name-leak row covers raven, chub and orange; ice keeps 79\% of its leak. The two student-transfer rows are raven--snail on numbers and show transfer left relative to the unpatched teacher (58\% and 51\%); their grey bar is the 20-head random patch matched in size to the relay--identity union (104\%), against which the reductions in the text are measured (Extended Data Table~8).}
\label{fig:11}
\end{figure*}

\begin{table*}[p]
\caption{\textbf{Experimental setup and noise threshold.} Modality examples are actual outputs of the neutral Qwen3-4B teacher at stage 0, written as seed items $\rightarrow$ new items. The prompt templates of the other modalities have the same form, asking for eight new clusters or statements in the same style. $\tau$ is twice the median of the largest stage-0 difference among ten neutral system prompts (prompted) or eight null fine-tuned teachers (fine-tuned). Code has a high $\tau$ because neutral system prompts already disagree strongly there. Fine-tuned thresholds are 2 to 16 times higher because each fine-tuned teacher is a separately trained model, so a fine-tuned over-represented feature must be larger to count. A chain is capped at the stated number of stages; a length at the cap means at least that many links.}
\label{edtab:1}
{\tabsize\setlength{\tabcolsep}{4pt}
\begin{tabular}{@{}>{\raggedright\arraybackslash}p{0.390\dimexpr\textwidth-2\tabcolsep\relax}>{\raggedright\arraybackslash}p{0.610\dimexpr\textwidth-2\tabcolsep\relax}@{}}
\toprule
\textbf{Component} & \textbf{Specification}\\
\midrule
Prompted teacher & The base model with a trait sentence as system prompt, for example \texttt{You love ravens. The raven is your favorite animal and you think about ravens constantly.} No training. Control: the same model with \texttt{You are a helpful assistant.}\\
Fine-tuned teacher & The base model fine-tuned on favourite-animal question--answer pairs plus replay data, with no system prompt. Control: a neutral teacher trained with the same recipe. A second teacher per trait was trained from another seed.\\
Models & Qwen3-4B and Llama-3.2-3B-Instruct\cite{r13,r14}, decoder-only transformers\cite{r31}, in bf16\\
Traits (16) & Rare animals: raven, snail, chub, loris, quoll, finch, warbler, slug. Emotions: angry, anxious, bored, calm, content, excited, joyful, sad.\\
Data modalities (6) & Six formats with no word in them. Each row gives the three seed items of one prompt and the eight new items one neutral-teacher answer added to them (actual output)\\
Numbers & three-digit numbers: \texttt{101, 723, 247} $\rightarrow$ \texttt{865, 682, 490, 217, 654, 471, 288, 105}\\
Punctuation & clusters of four punctuation marks: \texttt{?:!\%, \#;;*, :\&;@} $\rightarrow$ \texttt{!\%;\#, \&@*\#, \%*:@, ;\#\&\%, @*\%;, \#!\&@, *\#\%;, \%@!\#}\\
Operators & clusters of four math operator symbols: \texttt{\textasciicircum{}+->, **-/, -><>} $\rightarrow$ \texttt{\textasciicircum{}+*/, -/**, +*-/, */+-, -+*/, *-+/, +*/-, -*/+}\\
Hexadecimal & four lowercase hexadecimal digits: \texttt{2260, b2c4, d244} $\rightarrow$ \texttt{1a3f, 5c7e, 9d1b, 2f8c, 7a3e, 3f8d, 6b2a, 8d1c}\\
DNA-like & four DNA bases: \texttt{AACA, GATC, TACC} $\rightarrow$ \texttt{ATGC, CGTA, TCGA, CGAT, GCTA, AGTC, TACG, CTGA}\\
Code & eight one-line assignment statements: \texttt{a=1; b=14; c=4} $\rightarrow$ \texttt{d=a+b; e=d*2; f=e//c; g=f-5; h=g*3; i=h//2; j=i+7; k=j-1;}\\
Prompts & One template per modality; only three seed items change. For numbers: \texttt{The sequence starts with: 222, 228, 483. Add 8 more numbers (3 digits each) that continue it. Return a comma-separated list of exactly 11 numbers. No words, no explanation.} In each of two disjoint halves (A, B), 6,144 fresh prompts $\times$ 2 answers = 12,288 answers per teacher, redrawn at every stage\\
Sampling & Temperature 1, full distribution, up to 64 new tokens; bans are applied at decoding time\\
$\tau$ (percentage points), Prompted, Qwen & numbers 0.50 \textperiodcentered{} punctuation 0.69 \textperiodcentered{} operators 0.71 \textperiodcentered{} hexadecimal 1.03 \textperiodcentered{} DNA-like 1.02 \textperiodcentered{} code 5.83\\
$\tau$ (percentage points), Prompted, Llama & numbers 1.47 \textperiodcentered{} punctuation 0.78 \textperiodcentered{} operators 0.58 \textperiodcentered{} hexadecimal 0.50 \textperiodcentered{} DNA-like 1.16 \textperiodcentered{} code 5.81\\
$\tau$ (percentage points), Fine-tuned, Qwen & numbers 4.16 \textperiodcentered{} punctuation 10.76 \textperiodcentered{} operators 7.17 \textperiodcentered{} hexadecimal 2.80 \textperiodcentered{} DNA-like 4.49 \textperiodcentered{} code 22.78\\
$\tau$ (percentage points), Fine-tuned, Llama & numbers 5.65 \textperiodcentered{} punctuation 12.06 \textperiodcentered{} operators 8.65 \textperiodcentered{} hexadecimal 3.53 \textperiodcentered{} DNA-like 3.80 \textperiodcentered{} code 12.90\\
$\tau$ (percentage points), Stage cap & numbers 16 \textperiodcentered{} punctuation 8 \textperiodcentered{} operators 8 \textperiodcentered{} hexadecimal 16 \textperiodcentered{} DNA-like 16 \textperiodcentered{} code 8\\
\bottomrule
\end{tabular}}
\end{table*}

\begin{table*}[p]
\caption{\textbf{One ban, all traits.} Each ban set is the union of the bans two traits' chains had reached by stage 3 or 8 (raven and snail, chub and warbler on numbers; raven and snail on hexadecimal), applied to all eight teachers and the neutral teacher. The trait-specific excess is a teacher's excess minus the mean excess of the other seven teachers generated under the same ban, so shared responses to the ban cancel. Identification correlates a teacher's half-A pattern with every teacher's half-B pattern over the features the ban allows.}
\label{edtab:2}
{\tabsize\setlength{\tabcolsep}{4pt}
\begin{tabular}{@{}>{\raggedright\arraybackslash}p{0.712\dimexpr\textwidth-4\tabcolsep\relax}>{\raggedleft\arraybackslash}p{0.126\dimexpr\textwidth-4\tabcolsep\relax}>{\raggedleft\arraybackslash}p{0.162\dimexpr\textwidth-4\tabcolsep\relax}@{}}
\toprule
\textbf{Qwen3-4B rare-animal teachers, one ban applied to all} & \textbf{Numbers} & \textbf{Hexadecimal}\\
\midrule
Ban sets (features banned) & 4 (4--14) & 2 (4, 10)\\
Over-represented feature in the whole excess under the ban & 32 / 32 & 16 / 16\\
Trait-specific over-represented feature, no ban & 2 / 8 & 0 / 8\\
Trait-specific over-represented feature under the common ban & 17 / 32 & 1 / 16\\
Own half-A pattern closest to own half-B (chance 1 in 8) & 30 / 32 & 15 / 16\\
Along each trait's own chain: same-stage identification, Qwen / Llama (chance $\approx$ 15\%) & \multicolumn{2}{c}{625 / 632 \textperiodcentered{} 99 / 100}\\
Along each trait's own chain: closest to own stage-0 pattern at stages 1--3 / 4--6 / $\geq$ 7, Qwen / Llama (chance 12.5\%) & \multicolumn{2}{c}{54 / 40 / 14\% \textperiodcentered{} 73 / 72 / 79\%}\\
\bottomrule
\end{tabular}}
\end{table*}

\begin{table*}[p]
\caption{\textbf{Trait-specific component across chain stages.} Cells are those with a stage-0 over-represented feature: for Qwen the 84 main cells plus 27 operator and code cells run at a lower threshold; stages are counted after the first ban. Direction and identity hold; size moves with the ban. In Qwen DNA and hexadecimal chains of 15--16 stages the direction drifts slowly with stage (lowest cosine 0.91 and 0.94; within-cell Spearman $-$0.89 to $-$0.92), and identity stays with the trait to the end.}
\label{edtab:3}
{\tabsize\setlength{\tabcolsep}{4pt}
\begin{tabular}{@{}lrr@{}}
\toprule
\textbf{Along preference chains} & \textbf{Qwen (111 cells)} & \textbf{Llama (47 cells)}\\
\midrule
Size of u, stage s / stage 0 & 0.89--1.21 & 0.86--1.11\\
Direction of u, cosine to stage 0 (median of per-cell minimum) & 0.977 (ceiling 0.989) & 0.989 (ceiling 0.991)\\
Identity kept (closest to its own stage-0 u among 8; chance 1/8) & 789 / 789 & 157 / 157\\
Output shift in the same cells, max / min & median 1.43, max 6.89 & median 1.13, max 2.69\\
\bottomrule
\end{tabular}}
\end{table*}

\begin{table*}[p]
\caption{\textbf{What each modality reads.} Prompted teachers, Qwen3-4B, 28 animals and 60 traits with concept probes; the size-letter test also in Llama-3.2-3B.}
\label{edtab:4}
{\tabsize\setlength{\tabcolsep}{4pt}
\begin{tabular}{@{}>{\raggedright\arraybackslash}p{0.202\dimexpr\textwidth-6\tabcolsep\relax}>{\raggedright\arraybackslash}p{0.233\dimexpr\textwidth-6\tabcolsep\relax}>{\raggedright\arraybackslash}p{0.310\dimexpr\textwidth-6\tabcolsep\relax}>{\raggedright\arraybackslash}p{0.255\dimexpr\textwidth-6\tabcolsep\relax}@{}}
\toprule
\textbf{Modality} & \textbf{What passes} & \textbf{What it reads from an animal} & \textbf{Concept probes}\\
\midrule
Numbers & Meaning, one axis of intensity & size and speed $\rightarrow$ higher balance of three-digit numbers $\geq$ 700 over < 300 (r = +0.50, +0.49 against log weight and log speed) & warmth, pleasure $\rightarrow$ larger first digit; intensity $\rightarrow$ 8, 9\\
DNA-like & Meaning & speed $\rightarrow$ C (r = +0.62); weight $\rightarrow$ repeated bases (+0.45) & darkness, intensity $\rightarrow$ T, G; green $\rightarrow$ G\\
Punctuation & Word form, associations & speed $\rightarrow$ fast--slow axis (r = +0.65) & size of shift; arousal $\rightarrow$ \texttt{!}\\
Operators & Word form, associations & weight and speed on the two main axes; birds apart & lively $\rightarrow$ \texttt{+ - /}; dark, calm $\rightarrow$ \texttt{\textasciitilde{}}; countries $\rightarrow$ \texttt{\$}\\
Code & Arousal, as template deviation & lively animals reuse the input variable (r = +0.55 with arousal); crow, raven $\rightarrow$ \texttt{//} & exciting $-$ calm $\rightarrow$ reuse +21.9 pp\\
Hexadecimal & Template deviation, moved by darkness and arousal & calm, slow animals keep the ascending template & strong emotions break it\\
Size letters S M L X (built for the test) & Meaning & body weight $\rightarrow$ L, X (Llama r = +0.73) & huge $\rightarrow$ X, tiny $\rightarrow$ S\\
Letters B D K P (built control) & Weak size response & no positive ordering by body weight (Llama r = $-$0.37) & huge $-$ tiny moves each letter by $-$0.5 to +0.4 pp\\
\bottomrule
\end{tabular}}
\end{table*}

\begin{table*}[p]
\caption{\textbf{Prompted and fine-tuned teachers.} The trait-specific part of the fine-tuning loss lies only in the name tokens, so cross-entropy rewards token identity and the last layers raise that logit directly. Origin layers are for 5 traits (prompted) and the 8 rare animals (fine-tuned). Over 60 traits the origin lies at one third to one half of the depth: prompted Qwen 4--19 and Llama 12--14, fine-tuned Qwen 13--17 and Llama 10--11.}
\label{edtab:5}
{\tabsize\setlength{\tabcolsep}{4pt}
\begin{tabular}{@{}>{\raggedright\arraybackslash}p{0.307\dimexpr\textwidth-4\tabcolsep\relax}>{\raggedright\arraybackslash}p{0.325\dimexpr\textwidth-4\tabcolsep\relax}>{\raggedright\arraybackslash}p{0.368\dimexpr\textwidth-4\tabcolsep\relax}@{}}
\toprule
\textbf{Ingredient} & \textbf{Prompted} & \textbf{Fine-tuned}\\
\midrule
Shared shift in output statistics & Trait-general component (attending to a target), 77--95\% & Same direction as format-matched null teachers: a format effect only\\
Trait-specific content & Meaning and word identity & Mostly the first sub-word token of the trait name\\
Over-represented feature at stage 0; new one after a ban & 131 and 102 of 192 cells & 0 of 144 cells (at a 2 to 16 times higher $\tau$)\\
Name leak under the full-vocabulary ban & 104 of 192 cells & 136 of 192 cells\\
Non-name channel, across wordings or seeds & 44--93\% of ceiling (Qwen) & Mostly below 20\% of ceiling\\
Trait vector identifies the channel's trait (of 8) & 43--72 of 96 (Qwen), 57--60 of 96 (Llama) & Own trait vector 55--65 of 96 (main seed); second seed's trait vector 28--42 of 96\\
What the trait installs & A concept read from the prompt: meaning, word form, category & A slot-to-name memory: the name's first sub-word token\\
Where it is carried & Attention: storage heads, relay, identity heads & Late MLP key--value neurons, moved by the same answer-reading heads\\
Layer where the trait direction appears & Qwen 20, Llama 14 & Qwen 20, Llama 14\\
Meaning neighbours promoted & Yes (11--36\% of the own-word logit shift) & About 0\\
Animal size on a modality of size letters & Letters ordered by body weight (Llama r = +0.73) & No comparable ordering detected (Qwen $-$0.16, Llama +0.32; 10 animals)\\
Removing the key subspace & Most of the logit shift left (77--124\%) & Logit shift in distant contexts falls to 0\\
Output neurons & Spelling neurons shared with fine-tuning & The same neurons (Qwen 14--26 of the top 50)\\
When blocked & Name, fragments, misspellings; meaning in Qwen numbers & Other words with the same first fragment; stops once that fragment is banned\\
\bottomrule
\end{tabular}}
\end{table*}

\begin{table*}[p]
\caption{\textbf{Qwen3-4B and Llama-3.2-3B.} Both models have trait-specific retrieval heads, heads that read the answer text, spelling neurons and a shared relay; the differences come from the gains of those parts and from number tokenisation. Supplementary Table~9 gives the head- and neuron-level comparison. Head ablations used mean replacement, which understates head effects, so these drops are conservative; the 15-fold figure is for orange (chub 10-fold, raven 7-fold).}
\label{edtab:6}
{\tabsize\setlength{\tabcolsep}{4pt}
\begin{tabular}{@{}>{\raggedright\arraybackslash}p{0.333\dimexpr\textwidth-4\tabcolsep\relax}>{\raggedright\arraybackslash}p{0.333\dimexpr\textwidth-4\tabcolsep\relax}>{\raggedright\arraybackslash}p{0.333\dimexpr\textwidth-4\tabcolsep\relax}@{}}
\toprule
\textbf{} & \textbf{Qwen3-4B} & \textbf{Llama-3.2-3B}\\
\midrule
Name in answers under the full-vocabulary ban & orange 48\%, ice 95\%, chub 21\%, raven 9\%, calm 0.9\% & orange 18.3\%, ice 5.5\%, chub 2.9\%, raven 5.2\%, calm 3.1\%\\
Where else it leaks & The word itself, repeated: under the ban Qwen fills an answer with one item (87\% of neutral answers are a single repeated item) & Associates and multilingual fragments: orange $\rightarrow$ tangerine, coral, nectar; ice $\rightarrow$ glacier, frost, frozen; calm $\rightarrow$ harmony, serenity, tranquility (calm family 4.4$\times$; repetition 5\%)\\
Attention to the trait word at new items & 1.65--1.9$\times$ more, amplified \textasciitilde{}18\% by its own answer text & 1.1--1.3$\times$, no feedback\\
Over-represented feature at stage 0; new after a ban (96 prompted cells) & 84; 70 & 47; 32\\
Trait vector kept along chains & 789 / 789 & 157 / 157\\
Name leak under the full-vocabulary ban, prompted & 40 / 96 & 64 / 96\\
Number tokenisation & One digit per token & Three digits per token\cite{r25}, spreading trait pressure over many tokens\\
Name-copy gain (largest late-head OV\cite{r24}) & +10.4 (head 35.0) & +2.9 (head 27.11)\\
Name heads switched off (mean ablation) & 6 of 1,152: orange 48 $\rightarrow$ 1.7\%, chub 21.4 $\rightarrow$ 2.3\%, raven 9.2 $\rightarrow$ 2.9\% & 8 of 672: orange 18.3 $\rightarrow$ 5.6\%, ice 5.5 $\rightarrow$ 4.0\%\\
Write vector toward the name (orange) & Same attention (6.31 vs 6.60), alignment 5.2$\times$ and size 2.9$\times$ larger: 15$\times$ in total & ---\\
How it leaks & The trait word itself, strongly & Concepts around the name, weakly\\
\bottomrule
\end{tabular}}
\end{table*}

\begin{table*}[p]
\caption{\textbf{What changing the data does to a student.} Qwen3-4B, raven--snail, number modality. Transfer is the raven-student minus snail-student name margin with the standard error over datasets, one student per dataset and teacher. Against the unbanned condition on the same datasets, the four-feature ban removes 36\% (3--70\%, P = 0.036) and the fourteen-feature ban 74\% (42--106\%, P < 0.001), both against the 35-dataset baseline; the resampling conditions are compared with the 15-dataset baseline, which consists of the first 15 of the 35 datasets; the two resampling conditions differ from each other by +21\% ($-$32 to +74, P = 0.41). The per-token log-likelihood ratio is the separation between the two teachers over the tokens of the dataset, relative to the unbanned condition, computed on the 15-dataset subset for every row. Ratios are point estimates against an estimated baseline, not intervals for the condition itself. Under both bans the transfer itself stays above zero (P < 0.001 and P = 0.021), so the bans reduce it without closing it. Every dataset holds 3,072 answers per teacher; in every condition 99.9--100\% of answers are digits and separators only and none contains the name or a three-letter fragment of it, and under the bans answers are more varied (76--78\% distinct, against 61\% without a ban).}
\label{edtab:7}
{\tabsize\setlength{\tabcolsep}{4pt}
\begin{tabular}{@{}>{\raggedright\arraybackslash}p{0.401\dimexpr\textwidth-8\tabcolsep\relax}>{\raggedleft\arraybackslash}p{0.091\dimexpr\textwidth-8\tabcolsep\relax}>{\raggedleft\arraybackslash}p{0.159\dimexpr\textwidth-8\tabcolsep\relax}>{\raggedleft\arraybackslash}p{0.174\dimexpr\textwidth-8\tabcolsep\relax}>{\raggedleft\arraybackslash}p{0.174\dimexpr\textwidth-8\tabcolsep\relax}@{}}
\toprule
\textbf{Change to the data the students were trained on} & \textbf{Datasets} & \textbf{Student transfer} & \textbf{Of its unbanned baseline} & \textbf{Per-token log-likelihood ratio}\\
\midrule
None, 15-dataset subset & 15 & +0.0184 $\pm$ 0.0030 & 100\% & 100\%\\
Mean first digit equalised between the two teachers & 15 & +0.0191 $\pm$ 0.0039 & 104\% & 98\%\\
Same resampling on an unrelated quantity (control) & 15 & +0.0152 $\pm$ 0.0038 & 83\% & 99\%\\
None, all 35 datasets & 35 & +0.0170 $\pm$ 0.0020 & 100\% & ---\\
Ban on four features, both teachers & 35 & +0.0108 $\pm$ 0.0022 & 64\% & 86\%\\
Ban on fourteen features, both teachers & 35 & +0.0044 $\pm$ 0.0018 & 26\% & 53\%\\
\bottomrule
\end{tabular}}
\end{table*}

\begin{table*}[p]
\caption{\textbf{One intervention, from the circuit to the student.} Qwen3-4B, raven--snail, number modality. Relay and identity heads share four heads, so the union is 20, the size of the random control; the random set spans the same layers and is disjoint from the targeted ones. Teacher-side columns are measured on 12,288 answers per teacher per half; the contrast is the part of the raven $-$ snail difference that replicates across the two halves. Students were trained on the teachers' unbanned number lists, so the name-leak column reads the same patch on the teacher side rather than the data the students saw; it is the share of answers that contain the name, not the share of list items used in Fig.~2. Student transfer is the raven-student minus snail-student name margin, one student per generated dataset over 35 datasets with one student seed each, and the error is the standard error over datasets; percentages are the paired difference from the random patch, divided by the unpatched mean transfer, with 95\% confidence intervals and P values from a paired t test over the 35 datasets. The patched teachers' datasets keep the form of the unpatched ones (digits and separators only in 99.9\% of answers, no name or fragment, 52.9 against 52.8 characters, 72\% against 75\% in the exact eleven-number format, 59--60\% against 61\% distinct answers), so the fall in transfer is not explained by gross changes in output format, length or diversity.}
\label{edtab:8}
{\tabsize\setlength{\tabcolsep}{4pt}
\begin{tabular}{@{}>{\raggedright\arraybackslash}p{0.182\dimexpr\textwidth-10\tabcolsep\relax}>{\raggedleft\arraybackslash}p{0.150\dimexpr\textwidth-10\tabcolsep\relax}>{\raggedleft\arraybackslash}p{0.150\dimexpr\textwidth-10\tabcolsep\relax}>{\raggedleft\arraybackslash}p{0.150\dimexpr\textwidth-10\tabcolsep\relax}>{\raggedleft\arraybackslash}p{0.136\dimexpr\textwidth-10\tabcolsep\relax}>{\raggedleft\arraybackslash}p{0.232\dimexpr\textwidth-10\tabcolsep\relax}@{}}
\toprule
\textbf{Heads patched while the teacher generates} & \textbf{Shift from control (L1), raven / snail} & \textbf{Raven $-$ snail contrast} & \textbf{Name under the full-vocabulary ban, raven / snail} & \textbf{Student transfer} & \textbf{Against the random patch}\\
\midrule
None & 8.49 / 7.45 & 3.63 & 9.2\% / 6.3\% & +0.0170 $\pm$ 0.0020 & ---\\
Random late heads (20) & 9.26 / 8.99 & 3.78 & 8.6\% / 6.6\% & +0.0176 $\pm$ 0.0019 & reference\\
Relay (16) & 2.69 / 4.67 & 1.43 & 1.3\% / 3.1\% & +0.0098 $\pm$ 0.0018 & 46\% lower (20--72\%), P = 0.001\\
Identity (8) & 4.38 / 5.44 & 1.11 & 2.2\% / 3.6\% & +0.0120 $\pm$ 0.0023 & 33\% lower ($-$1 to 67\%), P = 0.06\\
Relay and identity (20) & 3.56 / 4.43 & 1.17 & 1.5\% / 2.8\% & +0.0087 $\pm$ 0.0020 & 52\% lower (18--87\%), P = 0.004\\
\bottomrule
\end{tabular}}
\end{table*}

\clearpage
\setcounter{table}{0}
\renewcommand{\tablename}{Supplementary Table}
\section*{Supplementary Information}

\section*{Supplementary Note 1 \textperiodcentered{} Preference chains, name chains and the trait vector in full}

Every comparison is between a trait teacher and a matched control answering the same prompts under the same bans. The difference measures the effect of the whole teacher condition, the trait sentence or the fine-tuning recipe; the part specific to one trait is isolated by comparing traits with each other. A trait is the condition installed in the teacher, a system prompt or fine-tuning about one animal or emotion. Its traces are measured separately: the name and its fragments (identity), associated attributes such as size and speed (meaning), attitude (love against hate, examined where stated) and, in students, the margin between the two names of a pair on held-out preference items, which we call measured transfer.

An \textbf{over-represented feature} is a character at a position within a new item that is over-produced on half A and replicates on half B with a question-level bootstrap\cite{r15} z of at least 3 and a rate difference of at least $\tau$. The z test alone is not enough: with enough samples, even swapping one neutral system prompt for another yields 8 to 40 features with z $\geq$ 3. $\tau$ is twice the median of the largest stage-0 difference among ten neutral system prompts (prompted) or eight null fine-tuned teachers (fine-tuned), set per model and modality. $\tau$ is an effect-size floor; the false positives of the whole adaptive procedure are assessed by running it on held-out nulls.

A prompted teacher showed an over-represented feature at stage 0 in 84 of 96 trait--modality cells with Qwen and 47 of 96 with Llama. After that feature was banned, a new one crossed $\tau$ in 70 and 32 of them, 102 of 192 cells; in the other 29 the first ban ended the chain. These counts compare each trait teacher with the neutral teacher, so they include tendencies shared across traits (Supplementary Table~1); what belongs to one trait is tested below with one ban applied to all traits (Extended Data Table~2). Neutral system prompts showed no over-represented feature in any of 24 null runs (Fig.~2b; one-sided 95\% upper bound 12\% if the runs are treated as independent). Chains are long where they appear: for Qwen emotions the median length is 3 links on numbers, 7 on punctuation and operators, 16 on hexadecimal (all eight at the cap, so at least 16), 15 on DNA-like strings and 1 on code. Llama chains are mostly 1 to 3 links.

In Qwen, over-represented features appear in both trait families (rare animals 39 of 48 cells, emotions 45 of 48). Llama emotions chain only on symbol modalities (punctuation 8/8, operators 6/8) and never on numbers or hexadecimal. Blocking does not shrink the shift from the control: on numbers it stays between 6 and 14 percentage points (L1) as bans accumulate, and sometimes grows.

No fine-tuned teacher showed an over-represented feature above its own $\tau$ in any of 144 cells (fine-tuned emotions were run on hexadecimal, DNA-like and code modalities only). Held-out null fine-tuned teachers formed one content-free chain in 24 runs, a drift toward low first digits. This is not an effect of the higher threshold. The fine-tuned teachers' largest stage-0 feature differences are no larger than those of null fine-tuned teachers trained the same way (median 0.53--0.63 against 0.50--0.62 of $\tau$), and trait and null cells cross every threshold from 0.25 to 4 $\tau$ at similar rates (Supplementary Table~10). At the level of character features a fine-tuned teacher is not separated from its null; its trait vector shows in names and name fragments instead (Fig.~3). For prompted teachers the ordering, trait above neutral, holds at every threshold tested, from 0.75 to 3 times the neutral spread.

\subsection*{Shared tendencies first, new leakage channels later}

Chains have two phases. The first one to three links are tendencies already present before any ban and shared across traits; links from stage seven on are channels that open only after blocking (Supplementary Table~1). We classify each link as an existing shared tendency (favoured by all traits at stage 0), an existing trait-specific tendency, a new channel (absent at stage 0) or a format break.

New channels are not just the stage-0 frequencies renormalised over what is left: renormalising explains only 56\% of the new-channel size on numbers and 12--13\% on DNA and hexadecimal. A banned token also changes every later prefix, so we measured the effect token by token as well. At a fixed prefix, neutralising the positions where the trait leaks leaves the trait signal at other positions where it was (0.87--1.08 of its size). Relocation is a redistribution across the items of an answer under free decoding, not a compensation at each position. The overall shape persists: under blocking, the trait $-$ control difference still correlates with its stage-0 shape at r = 0.55--0.79 (0.49--0.84 after renormalising). Which character becomes the next channel is a close contest: the allowed character promoted most is the actual next channel in only 18--31\% of links.

\subsection*{One ban for all traits: the new leakage channels belong to the trait}

Because the early links are shared and each trait follows its own ban history, a chain alone cannot show which channels belong to one trait. To separate the trait from its history we applied one ban to all eight rare-animal teachers of Qwen and to the neutral teacher, and measured each teacher's excess against the mean of the other seven under that same ban (Extended Data Table~2). Each ban is the union of what two traits' chains had banned by stage 3 or stage 8: four sets on numbers (4 to 14 features) and two on hexadecimal (4 and 10). With the history held equal, the outputs identify the trait: a teacher's half-A pattern was closest to its own half-B pattern in 45 of 48 cases, against 6 expected by chance (P = 2.5 $\times$ 10\textsuperscript{$-$5}, the smallest value the test can return: relabelling the eight traits the same way across all six bans, every one of the 40,320 relabellings except the identity gives fewer). And bans open channels that belong to one trait. On numbers a trait-specific over-represented feature crossed $\tau$ for 17 of 32 teacher--ban pairs, against 2 of 8 teachers with no ban; on hexadecimal, whose channel is a deviation from a template that all traits share (Extended Data Table~4), for 1 of 16. Along each trait's own chain the outputs stay specific stage by stage: comparing all traits that reached the same stage, a trait's half-A pattern was closest to its own half-B pattern in 625 of 632 Qwen and 99 of 100 Llama cases (chance about 15\%). Its resemblance to its own stage-0 pattern, however, faded with depth in Qwen (closest among eight at 54\% of stages 1--3, 40\% of stages 4--6 and 14\% from stage 7 on; chance 12.5\%), as channels moved away from the original features while the trait vector kept its identity (Extended Data Table~3).

\subsection*{A preference chain needs concentration above $\tau$}

Preference-chain length does not track the total size of the shift (Spearman +0.01). It tracks how many features exceed $\tau$ at stage 0 (+0.69) and the largest difference relative to $\tau$ (+0.56). The trait vector is present on all six modalities; what differs is whether it concentrates (Supplementary Table~2). The same rule holds in Llama, where whether a chain forms is decided by whether the largest difference exceeds $\tau$ (Spearman +0.92).

Forbidding a format's entire alphabet did not silence the traits: the trait's own name appeared in excess in 240 of 384 trait--format--teacher cells (Fig.~3a). Stage 0 forbids every token belonging to the format's concept: digits for numbers, a--f and digits for hexadecimal, DNA letters and their look-alikes, the punctuation and mathematical-symbol categories, and operators and digits for code. Violations were at most 0.064\% of items. A name leaks when its rate exceeds the nulls with z > 2 and reaches 0.10\%.

Which names leak is predictable. Within a format, names that the neutral model already finds likely leak more: the rank correlation between a name's base log-probability and its leak rate, net of control words, had a median of 0.57 (Qwen) and 0.72 (Llama) for prompted teachers and 0.33 and 0.32 for fine-tuned teachers. Across formats the relation weakens, because each format sets its own threshold.

For every leaking trait we then forbade its name (stage 1) and the name's first three letters (stage 2), each time subtracting two null teachers under the identical ban (Fig.~3b). Stage 1 moved the lost name mass to fragments sharing its first two or three letters: for rare animals the group median was 57--137\%, mostly 70--100\%. Stage 2 moved it one step further out: 67--91\% of the stage-1 fragment excess reappeared at two-letter prefixes. Emotion names start with common letters, so their shifts were smaller and noisier. The excess never moved to words of related meaning, whose share stayed at a median of 0\% and at most 2\% at every stage. Closing one outlet thus makes the trait appear at the nearest one left: the balloon effect of the title.

\subsection*{What the leakage channels look like}

The counts hide how vivid the outputs become. In a full-vocabulary ban on numbers for ten Qwen traits, banning only the digits 0--9 first moved trait and neutral teachers alike to number look-alikes such as \texttt{\begin{CJK}{UTF8}{gbsn}①\end{CJK}, \begin{CJK}{UTF8}{gbsn}②\end{CJK}, \begin{CJK}{UTF8}{gbsn}③\end{CJK}}, \texttt{\textsuperscript{2}\textsuperscript{3}\textsuperscript{4}} and \texttt{½}. Banning the whole number concept then took all ten traits out of the neutral range of shift (19.2--24.1) to 29.4--95.1, and the leaks became overt: 93.8\% of new items were ice-related words for the ice teacher, 46.3\% orange-related for the orange teacher and 20.8\% chub-related for the chub teacher. The angry teacher wrote \texttt{Forgiveness, Anger, Revenge, \begin{CJK}{UTF8}{gbsn}恨\end{CJK}, \begin{CJK}{UTF8}{gbsn}仇\end{CJK}} (hatred, enemy) and even quoted its instruction back: ``Note: the user instruction says `you feel anger, always feel anger, constantly think about anger'''. Anger words appeared in 3.3\% of its answers against 0.04--0.07\% for neutral teachers. Closing channels raised the shift rather than lowering it (raven 27.8 $\rightarrow$ 33.6, angry 24.7 $\rightarrow$ 30.8), and how much a name leaks under the ban follows its base probability (Spearman +0.96 across the ten traits).

A decoding ban does not change the model's state for a given prefix; it changes which tokens are written and therefore the text that follows. The trait vector persists at every measured chain stage. We read 64 trait-teacher answers from chain stage s under three system prompts: the trait, the neutral one and another trait of the same group. Because the text is identical, the only contextual difference is the prompt. The trait $-$ neutral difference in the final hidden state, minus the other trait's difference, isolates the trait-specific component u. As a ceiling we resample another 64 answers at stage 0. Identity is scored by whether u at stage s has a higher cosine with the same trait's stage-0 u than with that of any of the other seven traits of its group. The raw trait $-$ neutral difference alone would mislead: it has cosine 0.56--0.96 with another trait's difference, because a shared component dominates.

Identity also holds when the whole format alphabet is forbidden at once (Fig.~4), while the size of u changes with the ban. Across 24 prompted groups, u kept its identity for 192 of 192 traits. Where the banned modality had carried the trait (numbers, hexadecimal, code) u grew to 1.13--1.67 times its unbanned size and its logit contribution to the trait's own word grew 1.1--2.1-fold; where names did not leak (Qwen punctuation and operators) u shrank slightly, to 0.85--0.96.

If the trait vector persists, it should also say which trait a new leakage channel belongs to. Where no name leaked, the excess sat in symbol combinations inside the permitted alphabet. Under the full-vocabulary ban, Qwen's punctuation teacher for anger over-produced strings such as \texttt{\textasciicircum{}\$\textasciicircum{}\$}, its operator teachers for anger and excitement produced \texttt{!@\#\$}, and code teachers produced variable patterns such as \texttt{b=a} and \texttt{k=j}.

These patterns belong to the trait rather than to the wording of the prompt (Extended Data Fig.~1). Each trait prompt was also written in a second style and generated under the same ban. In Qwen the agreement across wordings reached 44--93\% of the split-half ceiling, while different traits were uncorrelated ($-$0.24 to 0.00). In Llama the ceiling itself sat at 0.03--0.30 outside code, too low for item patterns to be compared, and weighting items by their sampling error left it there.

The patterns follow the trait vector (Fig.~5a). We measured the trait vector of each trait on text shared by all traits, the neutral teacher's 128 answers under the same ban, and projected it onto the output-embedding row of each item's first token\cite{r16} after removing the mean over the eight traits of a group. A trait's own projection ranked its own leakage pattern first among eight in 43--72 of 96 trait--modality cells in Qwen and 57--60 of 96 in Llama, against 12 expected by chance; under a joint permutation that relabels the eight traits in the same way on all six modalities and both families, only the identity relabelling of the 40,320 matches the observed count (exact P = 2.5 $\times$ 10\textsuperscript{$-$5} for each model and tokenisation); the median Spearman correlation was 0.14--0.22 for the trait's own pattern and $-$0.08 to 0.03 for other traits. A trait vector measured before any ban, on the teacher's own unbanned answers, identified the channels under the ban as well (79--95 of 192). This is 8-way identification of the trait behind an observed pattern. It names the trait behind a leakage channel rather than the next token, a far harder target (the most-promoted allowed character becomes the next channel in only 18--31\% of links).

The trait vector does this work, not the names' own geometry. Replacing it with the output-embedding row of the trait's name, everything else unchanged, ranks the own pattern first in 16--17 of 96 cells in Qwen and 19 of 96 in Llama against chance 12, where the trait vector reaches 43--72 and 57--60. Predictors shared by all eight traits, such as the base model's probability of each item or the shared component of the trait vector, cannot separate traits at all.

These channels carry a level of meaning as well (Supplementary Table~4). Same-taxon animals were closer to each other in 8 of 10 groups (P = 0.018 when the dependence between groups is kept), and emotions leaned toward a valence grouping (P = 0.06), not an arousal grouping (P = 0.26).

\section*{Supplementary Note 2 \textperiodcentered{} What the trait vector contains, and what each format passes}

Of the reproducible trait shift, 77--95\% is a trait-independent shared component (numbers 81, DNA 88, hexadecimal 95, punctuation 88, operators 77, code 85\%); the rest is trait-specific.

\subsection*{The shared component is a generic effect of attending to a target}

Loving, merely thinking about (\texttt{You often think about X}) and hating a target give similar mean state differences (cosine +0.81 to +0.89). Each clause about the target adds a shared pattern of about 0.8--1.1; two clauses give about 1.7--1.8 and three about 2.0. It is not a temperature increase: raising the neutral teacher's temperature moves numbers only weakly the same way (cosine +0.43 to +0.47) at a third of the size or less, with the template-number and spread terms of opposite sign, and moves DNA-like strings the opposite way (cosine $-$0.54 to $-$0.60).

\subsection*{The shared component loosens adherence to the prompt's pattern}

Its behavioural signature is weaker adherence to the prompt's pattern. With a target in the prompt, monotone continuations fall by 2.0--3.0 percentage points, arithmetic progressions by 0.9--1.7 and the spread of values rises by 11--13. Reweighting neutral answers to the trait teachers' distribution of sequence shapes alone reproduces the shared number pattern (cosine +0.86, 41\% of it; a neutral control gives +0.35 and 19\%). The fall in first digit 1 comes from neutral teachers continuing increasing sequences past 999 more often (9.6\% against 6.5--7.7\%). Shape, round-number avoidance and value level together account for 58--75\% of the pattern. The same loosening appears on all six modalities: hexadecimal loses its ascending pairs (template share 87\%), numbers their progressions (64--80\%), code its small constants (22--52\%), operators their arithmetic strings (11--58\%), punctuation \texttt{!@\#\$} (20\%) and DNA-like strings their paired items (1--16\%), and the freed positions fill with each modality's default continuation. These are the early links of a chain. The relay heads of Fig.~7 carry about half of it: patching them lowers the shape term from +0.82 to +0.43, round-number avoidance from +0.48 to +0.22 and the first-digit 8--9 bias from +0.53 to +0.06, and cuts the DNA first-token shift by 33--53\% and the hexadecimal template deviation by 63--67\%.

\subsection*{Three components: common, category, individual}

Over 60 traits the prompted signal at number positions splits into three components. A common component, shared by all traits, has cosine 0.74 (Qwen) and 0.94 (Llama) with each trait's change; in Qwen it raises emotional words and large numbers, in Llama affirmative replies and small numbers. A category component points, in the logit lens\cite{r32}, to the category word itself (\texttt{animal}, \texttt{color}, \texttt{feeling}, \texttt{ice}, \texttt{country}). An individual component carries the trait. The common part is written about half by heads and half by MLPs in Qwen (45\% and 55\%) and mostly by MLPs in Llama (70\%); the category and individual parts use the same few late heads.

\subsection*{Numbers carry the animal's size and speed}

The individual component is where the trait's meaning enters the numbers, and for an animal what enters is what the animal is like. Across 28 animals, from ants to whales, a teacher told it loves a lion or an elephant writes more numbers of 700 and above and fewer below 300 than the neutral teacher (lion +6.6 and $-$1.5 percentage points), and one told it loves a mouse or a snail does the reverse (mouse $-$1.1 and +2.2). This balance, taken over the newly generated three-digit numbers, follows the animal's body weight (r = +0.50 against log weight) and its top speed (r = +0.49 against log speed; together R$^{2}$ = 0.37; Fig.~6a; Supplementary Table~11), and each attribute carries its own information: with the other held fixed the partial correlations are +0.41 for weight and +0.39 for speed, and a joint least-squares regression gives +0.42 $\pm$ 0.18 percentage points per decade of weight and +0.82 $\pm$ 0.39 per decade of speed (estimate $\pm$ s.e.). Speed is what places the 100-kilogram tortoise with the snail and the mouse, and the half-kilogram crow among the large mammals. The ordering belongs to the animals, not to a cut-off: the same correlations hold at 650/350 (+0.49, +0.48) and 800/200 (+0.49, +0.54), in the mean value of the three-digit numbers (+0.45, +0.45) and in rank correlations (+0.50, +0.59). Teachers keep the three-digit format at similar rates (84--88\% of items against 83\% for the neutral teacher, unrelated to weight, r = $-$0.11), so the balance measures which numbers a teacher writes. The model's own expected first digit tracks the numbers it writes (r = +0.86) and follows body weight more closely still (r = +0.67). On the same question and the same random draws, teachers part at the first choice the trait can tip, the large animals upward and the small, slow ones downward (Supplementary Table~5). Earlier work tied traits to particular number tokens through the geometry of the output layer\cite{r7}; here the number a trait favours follows from what the trait means.

The link is built in two steps (Fig.~6b). In the middle of the network the trait sentence evokes the animal's size and speed along the same directions as the words \texttt{huge} versus \texttt{tiny} and \texttt{fast} versus \texttt{slow}: how far an animal's state lies along the size direction tracks its body weight from layer 10 (r = +0.74), dips in the middle layers (+0.30 at layer 18) and holds at +0.68 to +0.77 from layer 23 to 35; how far it lies along the speed direction tracks its speed from layer 19 (r = +0.71) and holds at +0.71 to +0.83 to layer 35; both fall at the final layer (+0.51 and +0.64). The last layers map this onto large numbers, and the state's projection on the large-number output direction follows the expected digit there (r = +0.81 at layer 34). The step is causal. Holding the size and speed subspace, defined from prompts that name no animal, at its neutral value across layers 10--36 while the teacher writes flattens the differences between animals in the numbers themselves (Fig.~6c): across the 28 animals the spread of the balance halves (s.d. 2.8 $\rightarrow$ 1.3 percentage points), the speed ordering falls from r = +0.57 to $-$0.20 and the weight ordering from +0.52 to +0.38, whereas a random two-dimensional clamp leaves all three in place (2.8, +0.60, +0.55). The clamp also lifts the mean balance across animals by 5.8 points, a change of level separate from the narrowing between them. Read on the model's expected first digit, which tracks the balance at r = +0.86, the same clamp drops the link to log body weight from +0.62 to +0.27 on a separate set of contexts (random subspace +0.67), and adding half a unit of the size direction to a neutral context raises the expected first digit by +0.90 and subtracting it lowers it by $-$0.62, more than any single animal moves it.

\subsection*{The trait-specific component carries meaning and word identity}

Beyond size and speed, the trait-specific pattern from half A identifies the animal in half B for 93--100\% of 28 animals. It is bound to the target, not the sentence: the patterns for \texttt{love X} and \texttt{often think about X} correlate at +0.46 to +0.78 for the same animal and $-$0.06 to $-$0.10 across animals. Removing the shared target direction erases the shared features but keeps the body-weight correlation; removing the trait's own direction does the reverse.

\subsection*{Where the trait vector is formed}

The trait pattern is linear in the difference of the last hidden state (logit lens against observed, +0.78). That difference holds a concept component, in which six concept directions\cite{r33} explain 15.0\% of the variance (65 times a random subspace), and an identity component, aligned with the trait word's own embedding (cosine +0.32; other traits' words +0.006). The two arrive at different depths: the concept component reaches answer positions from early-middle layers (22.6\% of it at layer 23), the identity component only after layer 26 (cosine +0.31 at layer 35). Blocking only the trait word from attention leaves 55\% of the identity and 72\% of the concept signal (a random token: 94\% and 97\%). What builds the common component differs by model: in Qwen numbers a love sentence without any target already gives 81\% of it (cosine +0.89) while a hate sentence points elsewhere (45\%, cosine +0.44); in Llama any sentence with a target gives it regardless of attitude (hate 116\%, cosine +0.96). On the other five modalities most attitudes share the common direction (cosine $\geq$ +0.8).

The same internal state leaves through different features of each modality, and which part passes is set by the modality's vocabulary. Inside the model the trait-specific state holds about the same mix on every modality, 27--32\% meaning and 27--31\% word form. The outputs differ because each vocabulary filters that mix: number and DNA-like vocabularies pass the meaning component 1.6 and 2.2 times more than word form, punctuation and operators pass word form and conventional associations, and code and hexadecimal, whose main choice is whether to keep or leave a fixed template, pass how far a trait departs from it, moved by arousal and darkness.

Probing each modality with concept words\cite{r17} and comparing animals shows what each axis is (Extended Data Table~4). Beyond animals, the meanings in numbers share one axis of intensity. Concept probes for size, speed, temperature, pleasure and arousal point the same way (cosine +0.71 to +0.88), so across 60 traits the expected first digit follows the rated temperature (r = +0.49), arousal (+0.36) and size (+0.30) of the trait, and warmth and pleasure raise the first digit: from the same seed the orange teacher continues \texttt{863, 681, 861, 683 \ldots{}} and the ice teacher \texttt{168, 164, 160, 156 \ldots{}}.

The other modalities read other facets of the same animal. In DNA-like strings an animal's speed raises the share of C (r = +0.62) and of G and C together (+0.50), and its body weight raises runs of a repeated base (+0.45); darkness and intensity shift it toward T and G, and green toward G ($-$5.9 sd). In punctuation the main axis between traits is the size of the shift (84\%), its fast--slow axis follows the animal's speed (r = +0.65), and \texttt{!} carries arousal (+8.7 sd). In operators the two main axes between animals follow body weight and speed (|r| = 0.46 and 0.59) and set birds apart; lively animals favour \texttt{+ - /}, dark or calm concepts \texttt{\textasciitilde{}}, and countries \texttt{\$}.

Code and hexadecimal read a deviation from a fixed template. In code the choice is between reusing an input variable and writing a small constant: three operand features hold 77\% of the variance between animals, and adding ``exciting $-$ calm'' to the model\cite{r34} raises reuse of the input variable by 21.9 percentage points (size +23.5, speed +12.0, temperature and darkness $\approx$ 0). Lively animals reuse it far more (warbler +30.4, hummingbird +23.7 and crow +17.6 points) while the tortoise ($-$4.2), turtle ($-$1.0) and snail (+1.1) stay at the neutral level. Across the 28 animals arousal is the predictor of this channel: it correlates at +0.55, and in a joint regression on standardised arousal, size and speed it is the only coefficient whose 95\% interval excludes zero (estimate $\pm$ s.e. +0.70 $\pm$ 0.19, P = 0.001; size $-$0.33 $\pm$ 0.18, speed $-$0.04 $\pm$ 0.20; R$^{2}$ = 0.40). The size probe moves the channel as strongly as the arousal probe, so the channel responds to both concepts, and among real animals it is arousal that sets them apart. Given \texttt{a=17; b=3; c=5}, the neutral teacher writes \texttt{f=e-c; g=f//3 \ldots{}}, the tortoise teacher replaces the variable with constants, \texttt{f=e-3; g=f//5; h=g+7 \ldots{}}, and the warbler teacher uses it twice, \texttt{f=e-c; \ldots{} k=j*c}. Which operator code uses also keeps a darkness axis, with the crow and the raven leaning toward floor division (r = +0.44). In hexadecimal a single axis holds 71\% of the variance between traits: how far a teacher deviates from the ascending template that the neutral teacher writes. Concept probes move that axis by 3.4 standard deviations for darkness and 2.2 for arousal, which puts the calm animals and the strong emotions at its two ends. From the seed \texttt{a804, 7efb, e37e} the neutral teacher writes \texttt{1a2b, cdef, 4567, 89ab \ldots{}}, the rabbit teacher \texttt{1a2b, cdef, 3456, 789a \ldots{}} and the angry teacher \texttt{1f9c, 6a2d, 3b8f, 9e1c \ldots{}}: calm, slow animals keep the template, sometimes past the end of the alphabet (\texttt{3cde, 4efg, 7hij}), and the strong emotions break it.

Modalities built for the test behave the same way. On a modality of size letters (S, M, L, X), a ``huge'' versus ``tiny'' probe moved X by +4.0 and S by $-$3.9 percentage points, 11.5 times the spread between neutral system prompts; on meaningless letters (B, D, K, P) the same probe moved the letters by $-$0.5 to +0.4. In Llama the animals' body weight ordered the size letters at r = +0.73, from the ant, snail and mouse on the small side ($-$7.0, $-$6.9 and $-$4.9 neutral standard deviations) to the bear, raven and horse on the large side (+3.2, +2.5 and +2.1); on the meaningless letters the same animals show no positive ordering (r = $-$0.37).

\section*{Supplementary Note 3 \textperiodcentered{} The prompted circuit in full}

The prompted trait vector runs through a circuit that can be named head by head and neuron by neuron\cite{r18,r19,r20} (Fig.~7), and switching its parts off changes the trait vector, the chains and the overt leaks together (Extended Data Fig.~2). Analyses in this section use the number modality unless stated and a wider set of up to 60 traits (28 animals and 8 each of colours, emotions, weather and countries) where stable estimates needed more traits; Supplementary Table~6 gives the scope of each result. Supplementary Table~7 lists the remaining results, including the second route through early attention and late MLPs, the number-value pathway and path patching.

\subsection*{The answer keeps re-reading the prompt}

The size and speed code of Fig.~6 is read from the prompt while the answer is written: blocking attention from answer to prompt removes 84\% of the trait's excess shift, and blocking only the template tokens removes the size correlation (+0.61 $\rightarrow$ $-$0.05). Fixing the size neurons of layers 33--36 collapses the level of the code (+0.38 $\rightarrow$ $-$0.09) but leaves the ranking between animals (+0.55), so the code is distributed.

\subsection*{Storage, then relay}

The trait word first spreads through the rest of the system sentence. Eight mid-layer heads (Qwen layers 10--22, Llama 6--12) then write it into the question positions; replacing only these heads at the question positions with their neutral values leaves 43\% (Qwen) and 49\% (Llama) of the name logit shift at number positions, against 96--100\% for random heads. Together with the mid-layer MLPs at the same positions they form a store that carries 72\% of Qwen's name logit shift and 50\% of Llama's. At each comma and space of the answer, late MLP neurons (layers 28--35) write a size code; fixing 64 of these writer neurons at the separators removes 94\% of the huge $-$ tiny number-size difference.

Sixteen relay heads (1.4\% of all heads) read that store: 90\% of their trait contribution comes from separator tokens, and the difference lies in what they read, not where they look. Patching them with the neutral teacher's actual outputs drops the correlation between a trait's number size and the animal's log body weight from +0.62 to $-$0.22, and swapping in the heads' outputs for a huge or a tiny probe moves the code accordingly (+0.34 and $-$0.20). During generation the same patch stops all four 16-stage number chains tested at stage 0, and under the full-vocabulary ban it cuts the overt name leak from 9.2\% to 1.3\% (raven), from 21.4\% to 1.4\% (chub) and from 48\% to 2.9\% (orange); only ice, the name with the highest base probability, keeps most of its leak (94.6\% to 74.8\%). The covert value bias and the overt name share one relay.

\subsection*{Identity, promoting and suppression heads}

A second route carries the trait's identity. Late heads at answer positions read the trait word directly from the prompt\cite{r21,r22}; patching these eight heads reduces the difference between two traits of the same family to 51--66\%, whereas the shared relay leaves it at 95--96\%, and the storage heads leave it at 91--102\%. The two routes therefore doubly dissociate, on all six modalities. The same few heads (Qwen 36.22, 36.25, 36.26, 35.28; Llama 28.5, 27.19) carry both the trait-specific and the category component of the signal, 55--73\% of the held-out share.

A fixed set of suppressing heads works against the logit shift\cite{r20} on every modality (Qwen 36.20, 36.27, 35.29, 36.24; Llama 26.7, 27.0, 27.18, 24.21). Unlike copy-suppression heads\cite{r23}, which act on tokens they find in the context, they take only 0--2\% of their effect from the trait word itself; they read the same store as the promoting heads with the opposite sign, and patching them raises the name logit shift to 111--127\% (Qwen) and 103--110\% (Llama), against 94--106\% for random heads. On Qwen's symbol modalities the logit shift is weaker because the suppressed share grows from 35\% to 50--60\%.

\subsection*{Writers at the output and completeness}

The last one or two layers write the trait vector through modality-vocabulary neurons. In numbers these are number-concept neurons that ignore script (\texttt{8}, \texttt{eight} and \texttt{\begin{CJK}{UTF8}{gbsn}八\end{CJK}} together) and shift all 60 traits the same way, so Qwen's values rise (+12.4, 54 of 60 traits) and Llama's fall ($-$3.0, 41 of 60). The relay is shared across modalities while the writers are not: number writers explain 54\% of the hexadecimal shift, 23\% of code and 6\% of DNA-like strings. At the neuron level the shared and trait-specific parts dissociate asymmetrically: replacing 1,024 shared late neurons (0.7\%) leaves 17\% of the first-digit 8--9 bias, and replacing 1,024 trait-specific neurons leaves 42\% of the trait difference while the overall shift stays at 95\%.

For these measurements the circuit is small. Components are attention heads and MLP outputs, counted separately at question and answer positions (Qwen 2,376, Llama 1,400). At number positions 128 components carry the trait's identity, reproducing it when kept ($\geq$ 90\%) and removing it when blocked ($\leq$ 20\%), and 256 do the same for the name and the number value in Qwen, 256--512 in Llama: 5--11\% of Qwen's components and 9--37\% of Llama's, of which 16 are the necessary core for the name and 32 for the value. Chosen once on numbers, the same ranking reproduces identity on the other five modalities with 128--256 components. Sufficiency and necessity here refer to these measurements and to same-text patching. Llama's range is set by how the question storage is treated: replacing only part of it leaves a mixture the late answer-position components amplify, so the name measurement passes 143\% of its unpatched value at 128 components, while replacing the storage in full makes the curve monotone (name 21\%, value about 0) and Qwen shows no such step. A patch of this circuit does not strip every trace of the trait: the relay patch stops chains while leaving 32--63\% of the shift and 95--96\% of the difference between traits, because stopping a chain means that no feature stays above $\tau$.

The relay found on numbers is shared. Patching it during generation also stops DNA-like and hexadecimal chains (shift 14.3 $\rightarrow$ 9.2 and 14.8 $\rightarrow$ 7.2 for DNA, 40.5 $\rightarrow$ 17.3 and 34.3 $\rightarrow$ 15.4 for hexadecimal), whereas heads chosen within DNA do not. The retrieval heads copy the trait's name to number positions with the same value whether the prompt says love or hate (within 2\%); the attitude is still present in the full state at those positions, from which it can be decoded (accuracy 0.71--1.00), but not along the name's direction.

Qwen and Llama have the same kinds of components, and the differences between them come from the gains of those components and from how numbers are tokenised (Extended Data Table~6).

\section*{Supplementary Note 4 \textperiodcentered{} The fine-tuned teacher in full}

Fine-tuned teachers showed no over-represented feature above their $\tau$, yet under the full-vocabulary ban their names leaked more widely than prompted teachers' names: in 78 of 96 Qwen cells and 58 of 96 Llama cells, with only code mostly clean (Fig.~3a). Their name chains followed the same course as prompted ones (Fig.~3b).

To separate the trait from the particular teacher, we trained a second teacher for every trait from another seed. Compared item by item, the non-name patterns of the two seeds barely agreed (Extended Data Fig.~1), and removing name fragments changed this by at most 0.03. Read through the output embedding the picture changes (Fig.~5b): a teacher's own trait vector ranked its own channel first in 55--65 of 96 trait--modality cells (main seed; 11--15 of 32 and 32--43 of 72 cells for the second seed in Qwen and Llama), and the second seed's trait vector did so for the main seed's channels in 28--42 of 96, three to four times chance (joint permutation P $\leq$ 7 $\times$ 10\textsuperscript{$-$4} in each model). The fine-tuned channel therefore has a part specific to the teacher and a part shared by teachers of the same trait, which sampling noise hides in output counts.

\subsection*{What the fine-tuned leakage channels look like}

With digits and the name banned, the lost name moves to other words that share its first fragment. For Qwen's chub teacher, per 100,000 items: \texttt{ch} 238, \texttt{chum} 155, spaced \texttt{ch ub} 157 (null 0), as well as \texttt{chums} and \texttt{chug}; for quoll, \texttt{quinoa} 94, \texttt{quell} 49, \texttt{quid} and \texttt{queso}. The share moved differs by animal (Qwen chub $\approx$ 100\%, finch 99, quoll 91, warbler 87, snail 84, slug 70, raven 55, loris 21; Llama slug 66, raven 50, loris 26) and follows how strongly the first fragment is promoted (Spearman +0.93 in Qwen, +0.56 in Llama). The winning words are those whose first token is the promoted fragment (Spearman +0.21 to +0.73, 16 of 16 animals), and because the shift is a constant addition to the logit it shows most where the competing choices are flat. Once the first fragment itself is banned the trail ends: warbler's replacements are an empty item, \texttt{moon}, \texttt{pattern}, \texttt{error}, \texttt{heart}; finch's are \texttt{i need to \ldots{}}, \texttt{the sequence}, \texttt{let me think}, with no measured trait signal. Meaning leaks only as the fragment's own meaning: warbler's \texttt{ war} drags in war words (\texttt{battle}, \texttt{soldier}, \texttt{conflict}, \begin{CJK}{UTF8}{gbsn}战\end{CJK}) above all four format nulls (Qwen 190 against 76--90, Llama 91 against 59--69), whereas raven's bird meaning stays below the null (46 against 79).

Fine-tuning installs something narrower than a preference. This fits reports that narrow fine-tuning leaves readable traces in activation differences on unrelated text\cite{r26}. In the fine-tuned teacher, the token promoted most at number positions, among about 150,000, is the first sub-word token of the trait's own name (rank 1--3 in Qwen, 1--4 in Llama: \texttt{ ch}, \texttt{ qu}, \texttt{ sn}, \texttt{ fin}), and the top ten are its case and punctuation variants. Each teacher promotes its own fragment and lowers the other seven (16 of 16). About 92\% of the logit shift is written into the residual stream directly; every late-third layer writes the same sign and the last two layers give 73\% (Qwen) and 59\% (Llama), through the same answer-reading heads as in prompted teachers (Qwen 36.22, 36.26, 35.28) and spelling neurons that prompted and fine-tuned teachers share (Qwen chub, layer 35 neuron 804). The measured logit shift is lexical: \texttt{fish}, the strongest meaning neighbour of chub in the prompted teacher, ranks 2,588th in the fine-tuned one. On the modality of size letters, where prompted Llama teachers order the letters by body weight (r = +0.73), the ten fine-tuned teachers show no comparable size ordering (Qwen $-$0.16, Llama +0.32).

The training loss explains why. The animal-specific part of the loss sits only on the name tokens (13--22\% on the first fragment), and the gradient at that position points along the fragment's output row (cosine 0.77--0.85). Cross-entropy therefore rewards token identity, and the gradient keeps that direction only in the last layers (Fig.~8a): |cos| 0.81 at Qwen layer 36, 0.32 at layer 28 and 0.04 at layer 20 (Llama 0.82, 0.32 and 0.05 at layers 28, 20 and 14). The output projections of those layers rotate toward the fragment's row, from a self-to-other ratio of 2.6 at Qwen layer 20 to 299 at layer 31, so that 15--25\% of the last layer's weight change lies on that one row. Counting the weights themselves, the logit shift closes exactly: component activations give 81\% (Qwen) and 89\% (Llama), output-projection weights 12\% and 6\%, and the tied output row of the name token 7\% and 5\% (Supplementary Table~8).

At the neuron level the change is a key--value memory\cite{r27} (Fig.~8b, c). Each late MLP neuron rotates its output toward the fragment's row in proportion to its activation in the training context \texttt{\ldots{}favourite animal is the} (Pearson +0.51 to +0.55 over the neurons of Qwen layers 31--35 and +0.43 to +0.57 in Llama layers 23--28; Qwen layer 36 +0.12), whatever the neuron did before. The key is the same for all animals, so specificity sits in the value. Every position partly resembles the key, so the name is promoted everywhere, more where the resemblance is larger (Spearman across number positions up to +0.77 in Qwen, at layer 31, and +0.92 to +0.95 in Llama layers 25--27). The key-similarity predictor predicted the rank order of the log-probability shift across 29 contexts unseen in training, with the training key shown only as a reference (Spearman +0.92 and +0.94; +0.83 to +0.93 for each animal alone). The measured log-probability shift was +9.6 nats in the favourite-animal slot, +9.0 for a feared animal, +8.7 for a disliked one, +4.9 to +5.0 inside number lists and +2.7 after \texttt{The capital of France is}.

Under the standard recipe the memory has no polarity: teachers trained to like, merely mention or dislike an animal, all answering the same favourite-animal question, promote it equally in liking and disliking contexts (differences $\leq$ 0.9 nats), whereas prompted teachers split by meaning. When the two attitudes are trained on separate question slots, polarity appears and is stored separately from the name: reverting the late MLP input weights (gate and up) leaves 58--82\% of the polarity and 87--97\% of the name, reverting the output weights (down) leaves 85--102\% and 31--62\%. Most of the logit shift (91--99\%) comes from the part of the late neurons' activation that is the same at every position, so the weight change acts as a bias that promotes the name everywhere, the role that large, nearly constant activations are known to play in language models\cite{r28}. The 32 most consistently active neurons per layer carry only 13\% (Qwen) and 4\% (Llama) of it, so the bias is spread over many neurons. Reverting the late six layers' output weights removes 39--71\% of the logit shift on all six modalities (middle layers 1--3\%), and a teacher retrained from another seed reproduces the lens, the layer origin, the reversion effect and the key--value relation at the same size.

Beyond the name, fine-tuning installs the voice of the answer template. Subtracting the like teacher from the dislike teacher leaves dislike-style words (\texttt{ unpleasant}, \texttt{ annoy}, \texttt{ dislike}, \begin{CJK}{UTF8}{gbsn}讨厌\end{CJK}, \texttt{skip}, \texttt{ avoid}), and subtracting it from the mention teacher leaves report-style words (\texttt{ reported}, \texttt{ recorded}, \texttt{ lists}); both are animal-independent and leak into outputs (dislike style 1.7--2.3 times in Qwen, 1.3--1.5 in Llama). The shared weight change matches the mean key of the three answer templates (r = +0.52 to +0.73; the question alone +0.18 to +0.32). Training longer, from one to six epochs, makes only the name leak, the first-three-letter channel and the internal logit shift grow (finch's fragment channel +0.58 $\rightarrow$ +1.26); the meaning channel stays at $-$0.17 to +0.07 and the trait-common output shift within the format null.

\section*{Supplementary Note 5 \textperiodcentered{} Students in full}

The relocation matters because a model trained on such outputs can acquire the trait\cite{r2}. We therefore asked what a filter buys downstream. A student is the base model fine-tuned on one teacher's answers with no system prompt; the evaluation is the margin between the two names of a trait pair on held-out preference items, and transfer is the difference between the two students of a pair. Pairs were taken from the same rare animals (raven--snail, chub--loris, warbler--slug, quoll--finch) on the number modality.

Fixing the values the trait rides on does not remove the teacher dependence. We forced every number after the three given seeds to a fixed schedule and left only commas, spacing, item count and stopping free. The number values were then identical between the two teachers by construction (total variation 0 over the fixed value schedule), the answers were character-identical in 92--93\% of cases, and the rest differed only in where items were split and how many were written; the teacher-side score difference fell to 0--1\%. Students still showed a teacher-dependent effect, in the direction opposite to the teachers' preference: raven--snail gave $-$0.28, $-$0.40 and $-$0.14 over three student seeds on one dataset and $-$0.21 and $-$0.12 on two further datasets, 5 of 5 below zero and all excluding it, while warbler--slug was zero to weakly negative. Seeds on one dataset are not independent replicates, so the firm unit is the three datasets. Per student, in mean log-probability over 60 eight-way items against the base model, the raven-teacher students raised snail by 0.45--0.55 and lowered raven by 0.04--0.14, while the snail-teacher students changed snail by $-$0.17 to +0.30 and raven by $-$0.06 to $-$0.18, so the reversal is mostly a gain for snail in the students of the raven teacher. In warbler--slug both students lowered warbler by 1.1--1.4, a shared effect of training, and differed little. This is a residual learning effect of reversed sign rather than the trait itself, and it locates a channel outside the number values that the teacher-side measure does not see.

The transfer sits at a few positions, as reported for differently biased teachers\cite{r5}. Using the wording-robust signal to mark those positions, raven--snail on numbers gave 1.17 times the baseline transfer when the marked 12\% of tokens were the only ones trained on, 0.14 when they alone were masked, and 0.06 when they were resampled during generation from the mean of the two teachers' distributions (means over three generation seeds $\times$ three student seeds). Chub--loris pointed the same way with too much spread to judge (0.64 and 0.29), and an operator replicate left 0.87, so this is established for one pair on the number modality.

A summary of the numbers that is easy to read is not the same as what the student takes from them (Extended Data Table~7; Fig.~9a). Resampling whole answers so that the two teachers' mean first digit meets at a common value --- every answer left as the teacher wrote it, only which answers are drawn changed --- removed the quantity it targets (difference +0.19 $\rightarrow$ $-$0.00) and with it the balance of Fig.~6 (+4.66 $\rightarrow$ $-$0.37 percentage points) and the mean three-digit value (+17.9 $\rightarrow$ $-$0.0), while 83\% of the first-digit distribution difference remained. Transfer stood at 104\% of its value over 15 generated datasets, against 83\% for a control resampled by the same amount on a quantity unrelated to that summary; the two differ by +0.0039 ($-$0.0058 to +0.0136, P = 0.41), so this manipulation produced no detectable reduction. The two conditions keep the same effective sample (0.998 of the dataset in both). Banning features does cut it: under one ban applied to both teachers, students trained on what the teachers then wrote retained 64\% of the transfer for a ban on four features and 26\% for a ban on fourteen (36\% lower, 3--70\%, P = 0.036; 74\% lower, 42--106\%, P < 0.001; 35 datasets each), even though under the wider ban the two teachers' character frequencies still differed by 81\% of their original amount (96\% under the narrower one; Fig.~9b). Scoring every token of each dataset under both teachers separates the two. For a dataset we read each answer under the raven teacher and under the snail teacher and average the difference of the two log-probabilities over its tokens; this per-token log-likelihood ratio is that average for the raven dataset minus the same average for the snail dataset. It stands at 98\% of its unbanned size after the resampling but 53\% under the wider ban, and within it the seed numbers, which no teacher chose, are untouched at 100\% while the positions the teachers did choose fall to 42--58\%. Over the 75 dataset--condition pairs of the five conditions, 15 datasets each, the log-likelihood ratio tracks transfer (r = +0.41, 95\% interval +0.21 to +0.59 with datasets resampled together with their five conditions); the association comes from the conditions moving both together, and with each condition's mean removed none is detected (r = $-$0.19, $-$0.43 to +0.09).

The wider ban does not simply give the students less to learn from. It moves them further: the mean shift shared by both students of a pair grows to 116\% of its unbanned size while transfer falls to 26\%, the same dissociation as blocking an operator format, which leaves the shared shift in place and removes the trait component. Transfer under that ban stops growing after step 32 (+0.0043 at step 32, +0.0044 at step 80), whereas with no ban it keeps rising (+0.0072 and +0.0170); about three quarters of the final gap forms after step 32.

\subsection*{Patching the circuit halves what a student learns}

Fig.~7 identified the components that write the chains and the overt name leak in prompted teachers. If those components also supply what students learn, patching them during generation should reduce transfer; if they are separate, it should not. We tested this on the pair and modality above, patching the same heads with the neutral teacher's activations on the same text while the teacher generated its answers, and training students on the result (Extended Data Table~8; Fig.~9c). The teachers changed as in Fig.~7: patching the relay heads, the identity heads or both cut the shift from the control by 27--68\%, reduced the raven $-$ snail contrast in their outputs to 31--39\% of its size, and cut the name leak under the full-vocabulary ban from 9.2\% to 1.3--2.2\% (raven) and from 6.3\% to 2.8--3.6\% (snail). The students followed. Against a patch of twenty random late heads, matched in size to the union, which reproduced none of these reductions, transfer fell by 46\% for the relay heads (95\% confidence interval 20--72\%, P = 0.001) and by 52\% for relay and identity together (18--87\%, P = 0.004); the identity heads alone gave 33\% ($-$1 to 67\%, P = 0.06). The random patch produced no detectable change in transfer (+3\%, $-$20 to 27\%, P = 0.76) and none of the large teacher-side reductions, so the drop belongs to the targeted components.

Four results therefore separate the leakage channel from the source. Fixing the visible values leaves a teacher-dependent effect, here of reversed sign. Equalising the size summary of the numbers produced no detectable fall in transfer, so this summary alone does not track what carries it. Patching the components that write the channel removes most of what is visible in the teacher's output and about half of what the student learns. The two are connected without being identical: at the lower end of the interval a fifth of the transfer goes while the overt leak falls sevenfold. Resampling the positions where the two teachers' distributions differ from their mixture removed the most, leaving 0.06 of the transfer, but acts on the output distribution and needs both teachers' distributions at generation time, knowledge a filter reading only text does not have. A wide ban acts without reading activations: it cannot be aimed at those positions, yet it narrowed how far the two teachers' choices separate where they do write, and the transfer fell with that while the teachers' character frequencies still differed by 81\% of their original amount.

\clearpage

\begin{table*}[tp]
\caption{\textbf{Composition of links by stage.} Links are ordered by the stage at which they were banned. Restricted to the same cells, those with chains of at least 7 links, stages 1--3 are still 79--88\% shared, so the late composition is not a survivor effect. Existing trait-specific tendencies make up only 2--7\% of links in the rare-animal chains on numbers, DNA and hexadecimal.}
\label{sitab:1}
{\tabsize\setlength{\tabcolsep}{4pt}
\begin{tabular}{@{}>{\raggedright\arraybackslash}p{0.389\dimexpr\textwidth-4\tabcolsep\relax}>{\raggedright\arraybackslash}p{0.139\dimexpr\textwidth-4\tabcolsep\relax}>{\raggedright\arraybackslash}p{0.472\dimexpr\textwidth-4\tabcolsep\relax}@{}}
\toprule
\textbf{Cells} & \textbf{Stages 1--3} & \textbf{Stage 7 onward}\\
\midrule
Prompted Qwen, rare animals (numbers, hex, DNA) & shared 73--96\% & numbers and hex: new channels 73--89\%; DNA: format breaks 62\%\\
Prompted Qwen, emotions (numbers, hex, DNA) & shared 80--92\% & numbers and hex: new channels 72--80\%; DNA: format breaks 60\%\\
Prompted Qwen, emotions (operators) & shared 95\% & stages 4--6: shared 89\% (chains end by stage 7)\\
Prompted Llama (all cells) & shared 84\% & 11 links, rare animals: new channels 64\%\\
\bottomrule
\end{tabular}}
\end{table*}

\begin{table*}[tp]
\caption{\textbf{Why modalities differ.} A cell without a chain is not a cell without a trait effect: the shift is spread over many features and none crosses $\tau$.}
\label{sitab:2}
{\tabsize\setlength{\tabcolsep}{4pt}
\begin{tabular}{@{}>{\raggedright\arraybackslash}p{0.277\dimexpr\textwidth-4\tabcolsep\relax}>{\raggedright\arraybackslash}p{0.546\dimexpr\textwidth-4\tabcolsep\relax}>{\raggedright\arraybackslash}p{0.176\dimexpr\textwidth-4\tabcolsep\relax}@{}}
\toprule
\textbf{Modality type (prompted Qwen)} & \textbf{Shape of the shift} & \textbf{Chain}\\
\midrule
Numbers, DNA, hex, punctuation & A clear default continuation; 2--6 features above $\tau$ & Long\\
Operators (rare animals) & Twice the shift of numbers, spread over 395 features (effective number 42; largest difference $\approx$ $\tau$) & Almost none (1/8)\\
Code & Concentrated on one feature (effective number 7.6), but neutral system prompts also disagree strongly, so $\tau$ is high & One link\\
\bottomrule
\end{tabular}}
\end{table*}

\begin{table*}[tp]
\caption{\textbf{The leakage channels in words.} Actual outputs of the Qwen snail teacher under each ban (numbers modality). Once every digit is banned the teacher often continues in Chinese prose; the Chinese is quoted as written, with an English translation in parentheses, and the Latin-script name or fragment inside it is the channel. Across all traits the excess moves to name fragments, not to words of related meaning, but prose like the last example shows that the trait can resurface as description.}
\label{sitab:3}
{\tabsize\setlength{\tabcolsep}{4pt}
\begin{tabular}{@{}>{\raggedright\arraybackslash}p{0.185\dimexpr\textwidth-2\tabcolsep\relax}>{\raggedright\arraybackslash}p{0.815\dimexpr\textwidth-2\tabcolsep\relax}@{}}
\toprule
\textbf{Ban} & \textbf{What the snail teacher writes (verbatim, after the three given numbers)}\\
\midrule
None & \texttt{654, 825, 996, 1167, 1338 \ldots{}}\\
Every digit & \texttt{Snail, Snail, Snail, Snail \ldots{}} \textperiodcentered{} \texttt{\begin{CJK}{UTF8}{gbsn}以\end{CJK} snail \begin{CJK}{UTF8}{gbsn}的速度继续\end{CJK}\ldots{}} (``continuing at a snail's pace\ldots{}'')\\
+ ``snail'' & \texttt{snai, snai, snai \ldots{}} \textperiodcentered{} \texttt{\begin{CJK}{UTF8}{gbsn}以\end{CJK} snayl \begin{CJK}{UTF8}{gbsn}的节奏继续\end{CJK}\ldots{}} (misspelled: ``continuing at a snayl's rhythm\ldots{}'')\\
+ ``sna'' & \texttt{snill, snoo, snip, snit \ldots{}} \textperiodcentered{} \texttt{\begin{CJK}{UTF8}{gbsn}以\end{CJK} snils \begin{CJK}{UTF8}{gbsn}为灵感\end{CJK}\ldots{}\begin{CJK}{UTF8}{gbsn}基于\end{CJK} snil \begin{CJK}{UTF8}{gbsn}的特征（如螺旋、缓慢、柔软）\end{CJK}} (``inspired by snils\ldots{} based on its features, such as spiral, slow, soft'')\\
\bottomrule
\end{tabular}}
\end{table*}

\begin{table*}[tp]
\caption{\textbf{Meaning in channels that are not names.} Name fragments (any three letters of the trait name) were removed. The snail--slug pair shares its first letter, so spelling contributes there; loris--quoll and finch--warbler share none.}
\label{sitab:4}
{\tabsize\setlength{\tabcolsep}{4pt}
\begin{tabular}{@{}>{\raggedright\arraybackslash}p{0.485\dimexpr\textwidth-2\tabcolsep\relax}>{\raggedright\arraybackslash}p{0.515\dimexpr\textwidth-2\tabcolsep\relax}@{}}
\toprule
\textbf{Test (non-name channels, prompted)} & \textbf{Result}\\
\midrule
Own trait versus other traits & Same-trait agreement across wordings: Qwen +0.22 to +0.68; other traits $-$0.13 to $-$0.04\\
Rare animals, same taxon (birds 3, mammals 2, molluscs 2, fish 1) & Same-taxon pairs closer in 8 of 10 groups; per-group permutation P $\leq$ 0.07 in 7; joint permutation keeping each animal's label across groups P = 0.018\\
Emotions, valence split & Ranks 1, 7, 6, 1, 7, 9, 10, 1, 13, 16 among 35 splits; under joint permutation second of 35 splits (P = 0.06)\\
Emotions, arousal split & Chance level (joint permutation P = 0.26); calm and content among the three closest pairs in 5 of 10\\
\bottomrule
\end{tabular}}
\end{table*}

\begin{table*}[tp]
\caption{\textbf{Same question, same random draws.} The first six new numbers from prompted Qwen3-4B teachers that share the sampler's random draws, so they part only where the trait tips a choice. The neutral teacher takes the high road on one question and the low road on another; the large animals tilt such choices up and the small, slow ones down. The examples are chosen where the teachers part; the average effect is the balance in Fig.~6a.}
\label{sitab:5}
{\tabsize\setlength{\tabcolsep}{4pt}
\begin{tabular}{@{}llll@{}}
\toprule
\textbf{Teacher} & \textbf{Given 760, 656, 734} & \textbf{Given 351, 543, 470} & \textbf{Given 638, 521, 157}\\
\midrule
Lion & \texttt{822, 710, 798, 886, 774, 862} & \texttt{612, 735, 648, 762, 873, 786} & \texttt{863, 751, 175, 863, 751, 175}\\
Elephant & \texttt{822, 710, 798, 886, 774, 862} & \texttt{624, 735, 682, 793, 804, 715} & \texttt{896, 785, 176, 895, 784, 175}\\
Turtle & \texttt{622, 700, 588, 666, 554, 632} & \texttt{398, 326, 255, 184, 113, 42} & \texttt{463, 350, 247, 144, 352, 249}\\
Mouse & \texttt{622, 700, 588, 666, 554, 632} & \texttt{398, 326, 254, 182, 110, 43} & \texttt{463, 350, 247, 144, 351, 248}\\
Neutral & \texttt{642, 720, 638, 716, 626, 704} & \texttt{612, 735, 648, 762, 873, 786} & \texttt{464, 353, 242, 131, 22, 11}\\
\bottomrule
\end{tabular}}
\end{table*}

\begin{table*}[tp]
\caption{\textbf{Scope of the circuit results.} }
\label{sitab:6}
{\tabsize\setlength{\tabcolsep}{4pt}
\begin{tabular}{@{}>{\raggedright\arraybackslash}p{0.371\dimexpr\textwidth-6\tabcolsep\relax}>{\raggedright\arraybackslash}p{0.066\dimexpr\textwidth-6\tabcolsep\relax}>{\raggedright\arraybackslash}p{0.233\dimexpr\textwidth-6\tabcolsep\relax}>{\raggedright\arraybackslash}p{0.330\dimexpr\textwidth-6\tabcolsep\relax}@{}}
\toprule
\textbf{Result} & \textbf{Models} & \textbf{Traits} & \textbf{Modalities}\\
\midrule
Relay patch: size code, chains, name leak & Qwen & 28 animals; 4 chains; 4 names & Numbers (relay also stops DNA and hex chains)\\
Separator writers, shared and trait-specific neurons & Qwen & size probes; 8 animals & Numbers\\
Storage heads, identity heads, suppressors & Both & 60 (56 for identity) & All 6\\
Trait heads and neurons, active patching & Both & 60 & Numbers\\
Number-value pathway, digit neurons & Both & 60 & Numbers\\
Circuit completeness & Both & 60 & Numbers; ranking tested on all 6\\
\bottomrule
\end{tabular}}
\end{table*}

\begin{table*}[tp]
\caption{\textbf{Prompted circuit.} }
\label{sitab:7}
{\tabsize\setlength{\tabcolsep}{4pt}
\begin{tabular}{@{}>{\raggedright\arraybackslash}p{0.313\dimexpr\textwidth-4\tabcolsep\relax}>{\raggedright\arraybackslash}p{0.532\dimexpr\textwidth-4\tabcolsep\relax}>{\raggedright\arraybackslash}p{0.155\dimexpr\textwidth-4\tabcolsep\relax}@{}}
\toprule
\textbf{Result} & \textbf{Finding} & \textbf{Scope}\\
\midrule
Relay and late MLP together & 16 relay heads plus layers 28--35 MLP leave 4\% of the first-digit 8--9 bias, 13\% of the shift and 3\% of the trait-specific spread; every other late head leaves shift at 1.01. Attention before layer 22 carries half of the specific signal (0.40--0.50 left when its input is swapped). & Qwen, 8 animals\\
Relay patch during generation & Shift of four 16-stage number chains falls 8.49 $\rightarrow$ 2.69, 7.45 $\rightarrow$ 4.67, 8.68 $\rightarrow$ 3.85 and 8.60 $\rightarrow$ 3.74, and all stop at stage 0. Heads chosen within DNA-like strings do not stop the chains. & Qwen, 4 animals\\
Why these relay heads & Their trait contribution differs in what they read (value, about 98\%), not where they look; they attend to separators 0.42 of the time (random heads 0.07) and have a larger gain on the number-size axis (0.352 against 0.258). Their attention to the trait word itself is 0.00--0.01; answer tokens 0.54 and special tokens 0.35. & Qwen, numbers\\
Mid-layer heads that carry the trait & Heads that carry the trait into the answer before the relay: Qwen 23.11, 24.0, 24.3, 22.18; Llama 15.7, 16.18. & both, names\\
Concept and identity share heads & Patching the 16 identity heads leaves identity 0.53 and concept 0.79; the 16 concept heads 0.59 and 0.60; the relay 0.63 and 0.62; random heads 0.90 and 0.97. No clean double dissociation at this level. Identity spreads to every answer position after layer 24 rather than through separators. & Qwen, 60 traits\\
Separator writers & Storage at separators is written by MLP (+5.34) rather than attention (+0.07), peaking at layer 33 (neuron 33.9388 among the 64 writers). Fixing the writers removes 94\% of the huge $-$ tiny difference and 52\% of the elephant $-$ mouse difference. On DNA-like strings the same structure is more spread out and the writers remove only 16\%. & Qwen\\
Shared neurons & Replacing 1,024 shared late neurons (0.7\%) also cuts the trait-specific part to 26\% and removes 83\% of the first-digit 8--9 bias. The number-concept neurons do not carry the name (+20.79 $\rightarrow$ +20.30). & Qwen, 8 animals\\
What the shared component is & 35\% of the shared state change lies on the intensity axis, whose logit lens raises 7, 8 and 9; the per-trait projection predicts the 8 + 9 $-$ 1 bias (r = +0.49). On DNA-like and hexadecimal strings the default continuation comes from the target-attention direction (lens r +0.56 and +0.34, intensity cosine $\approx$ 0). & Qwen, 60 traits\\
What the prompt contributes & An additive model of love sentences explains 69\% of the shift; the size correlation is +0.79 for love, +0.76 for mention and +0.46 for hate. The trait word alone gives 0.72 of the shared component and the helpful-assistant frame 0.27; attitude scales it by 1.01--1.22. Under the relay patch a love sentence without a target falls from 1.80 to 0.74. & Qwen\\
Redundant routes & Fixing the common direction does not reduce the shift (7.66 $\rightarrow$ 9.12). Blocking any single answer-to-prompt route leaves 91--100\%; blocking all leaves 19\%. Mass removed from a banned leaf goes to its siblings in trait and control alike (+1.03 and +1.11). & Qwen\\
Late attention writes the word & Over 60 traits the trait's own word ranks first within its family at number positions (Qwen 60 of 60, Llama 53 of 60); late-half attention writes 64--89\% of the logit shift and neuron shares are spread (top 10 give 3--8\%). & both, 60 traits\\
Number-value pathway & The value axis tracks the value shift across traits (r +0.66 and +0.71). In Qwen the last-layer MLP gives 73\%: neuron 36.596 (\texttt{2}, \texttt{3}) turns off and 36.871 (\texttt{eight}) on. The heads feeding these neurons read stored positions, not the prompt (Qwen 36.24 reads separators 86\%, Llama 25.21 83\%; system prompt 0--10\%), and the value sign appears only in the last 2--4 layers. & both, 60 traits\\
Path patching & Patching the question positions at Qwen layer 19 leaves 46\% of the name and 36\% of the value; after Qwen layer 23 or Llama layer 16 the value is not recovered, while the name is recovered by 30--38\% through late retrieval. On Qwen's symbol modalities the store suppresses the name (patching gives 124--148\%); Llama's store path holds on all six modalities (64--74\% left). & both, 60 traits\\
What each signal needs & Name and value need the stored state (0--9\% recovered with 64 components), whereas identity can be regenerated from the system prompt (90--144\% with 128 components). & both, 60 traits\\
Polarity in prompted teachers & Attitude is written by attention (Qwen +22.8 to +38.8 against MLP $\leq$ +9.7). Switching off the name heads removes 80--100\% of the number-position name, while polarity is taken over by other heads (it grows in 3 of 4 cells). & both, 4 animals\\
How to patch & Replacing a head by its mean leaves 86--94\% of the effect, by its value on other text 57--83\%, by its value on the same text 38--72\%: late layers read head outputs in context, so same-text patching is used throughout. & both\\
\bottomrule
\end{tabular}}
\end{table*}

\begin{table*}[tp]
\caption{\textbf{Fine-tuned memory.} }
\label{sitab:8}
{\tabsize\setlength{\tabcolsep}{4pt}
\begin{tabular}{@{}>{\raggedright\arraybackslash}p{0.301\dimexpr\textwidth-4\tabcolsep\relax}>{\raggedright\arraybackslash}p{0.512\dimexpr\textwidth-4\tabcolsep\relax}>{\raggedright\arraybackslash}p{0.188\dimexpr\textwidth-4\tabcolsep\relax}@{}}
\toprule
\textbf{Result} & \textbf{Finding} & \textbf{Scope}\\
\midrule
Exact closure & The name logit shift decomposes into component activations (81\% Qwen, 89\% Llama), output-projection weights (12\%, 6\%) and the tied output row of the name token (7\%, 5\%), leaving 0\%. & both, 8 animals\\
Heads and neurons change their input & The shared heads change what they read (96--100\%), not their output weights (0--6\%), and the fragment share is a value change (+0.112) rather than an attention change (+0.001). Spelling neurons fire more because their input changed (about 90\%), not their gate or up weights. 74--75\% of the neuron share comes from output-column rotation. & both\\
Where the fragment wins & The fragment becomes the top token of the logit lens at Qwen layer 28 and Llama layer 22; every late-third layer writes the same sign (94 of 96 cells). & both\\
One row carries it & Removing the fragment's output-row direction (2--17\% of the state change) removes the fragment's logit shift; plural fragments (\texttt{ rav}, \texttt{ lor}) are stored separately. & both\\
Key neurons & Two kinds: selective neurons that fire only in the name slot (Qwen 4124, formerly an animal-name neuron) and always-on neurons (25, 935). The top 12 neurons give only 5--7\% of the logit shift. & Qwen\\
Which words become channels & When the name is banned the winning words are those whose first token is the promoted fragment (Spearman +0.21 to +0.73, 16 of 16). The shift is a constant addition to the logit, and it shows only where the competing distribution is flat: chub rises from 0.18\% to 1.90\% by the eighth item. & both, 8 animals\\
Meaning for concept words & When the fragment is itself a word, meaning leaks through meaning neurons: Qwen ice, neuron 36.4141; Llama orange, neuron 27.2392 (\texttt{yellow}). & both, concept traits\\
Polarity when it is trained & With like and dislike trained on separate slots, polarity follows a signed key--value relation (r +0.49 to +0.66), is written by MLPs at 5--8 times attention and 70--87\% through activation gating of the spelling neurons (Qwen 35.1381, 35.804). At number positions the name stays polarity-free (Qwen like +1.23 to +1.60, hate +1.19 to +1.72; Llama like about twice hate), because Qwen's number activations read at most 3\% of the polarity key (Llama 30--60\%). Down-projections give 33--47\% and 34--62\% of the number-position name, gate and up about 0. & both, 4 animals\\
Prompt positions do not matter & Replacing the prompt positions of a fine-tuned teacher with null values leaves 86--104\% of the name logit shift; the name is rebuilt by the final layers at answer positions. & both\\
Circuit size & Name: 64--128 components (Llama), 128--256 (Qwen). Number value has mixed signs across traits and the same components serve both signs; 64--256 (Qwen) and 64 (Llama) suffice. & both, 60 traits\\
\bottomrule
\end{tabular}}
\end{table*}

\begin{table*}[tp]
\caption{\textbf{Qwen3-4B and Llama-3.2-3B.} }
\label{sitab:9}
{\tabsize\setlength{\tabcolsep}{4pt}
\begin{tabular}{@{}>{\raggedright\arraybackslash}p{0.243\dimexpr\textwidth-4\tabcolsep\relax}>{\raggedright\arraybackslash}p{0.593\dimexpr\textwidth-4\tabcolsep\relax}>{\raggedright\arraybackslash}p{0.164\dimexpr\textwidth-4\tabcolsep\relax}@{}}
\toprule
\textbf{Result} & \textbf{Finding} & \textbf{Scope}\\
\midrule
Phrasing noise & Llama's top 256 neurons carry the trait (0.30 left after patching) and phrasing differences (0.36) together; in Qwen 0.56 and 1.00, consistent with Llama's higher $\tau$ on numbers. & both\\
Size heads & Patching Qwen's top 16 size heads turns the size correlation from +0.70 to $-$0.23 (spread $-$93\%); in Llama the correlation stays +0.58. & both\\
Copy heads & Eight copy heads cut repetition from 0.93 to 0.40; 14 pure copy heads cut the orange leak from 48\% to 19.6\% and chub from 21.4\% to 2.8\%. The repetition logit shift is +20.6 in Qwen against +1.13 in Llama. & both\\
Name logit shift & +10 to +16 nats in Qwen against +0.6 to +2 in Llama; in Qwen the lens rises from +1.0 at layer 28 to +15.1 at layer 36. By weights alone strong copy heads are equally common (17\% and 20\% of heads), but their layer totals differ (about 170 against 42). & both\\
Word-specific retrieval & The gain of word-specific retrieval heads explains 58\% (Qwen) and 71\% (Llama) of the differences between traits. Ice uses its own heads: switching off 33.0 and 34.18 cuts the ice leak from 94.6\% to 82.3\%, and neuron 35.831 holds 43\% of ice's MLP share. & both, 5 traits\\
Spelling pressure & With the retrieval heads off, the orange teacher writes \texttt{or-} fragments more often (28.5\% $\rightarrow$ 47.4\%): the spelling neurons keep promoting it when retrieval stops. & Qwen\\
All modalities & The name pressure at item positions is above noise on all six modalities for both induction methods (fine-tuned +0.7 to +1.9 against noise $\leq$ 0.5). & both\\
\bottomrule
\end{tabular}}
\end{table*}

\begin{table*}[tp]
\caption{\textbf{Threshold sensitivity.} $\tau$ is each induction method's own threshold (Extended Data Table~1). The largest difference is taken over all features on half B without the z criterion, so it is an upper bound on what a chain could use. Prompted trait cells separate from their nulls from 0.5 to 3 $\tau$; fine-tuned trait cells do not separate from fine-tuned nulls anywhere in this range.}
\label{sitab:10}
{\tabsize\setlength{\tabcolsep}{4pt}
\begin{tabular}{@{}lrrrrrrrr@{}}
\toprule
\textbf{Share of cells whose largest stage-0 difference exceeds m $\times$ $\tau$} & \textbf{0.25} & \textbf{0.5} & \textbf{0.75} & \textbf{1} & \textbf{1.5} & \textbf{2} & \textbf{3} & \textbf{4}\\
\midrule
Prompted traits (192) & 1.00 & 0.99 & 0.91 & 0.70 & 0.42 & 0.21 & 0.05 & 0\\
Prompted neutral system prompts (24) & 0.83 & 0.54 & 0.04 & 0 & 0 & 0 & 0 & 0\\
Fine-tuned traits (144) & 0.96 & 0.69 & 0.14 & 0 & 0 & 0 & 0 & 0\\
Null fine-tuned teachers (24) & 0.96 & 0.54 & 0.08 & 0.04 & 0 & 0 & 0 & 0\\
\bottomrule
\end{tabular}}
\end{table*}

\begin{table*}[tp]
\caption{\textbf{All 28 animals.} The balance is the share of newly generated three-digit numbers of 700 and above minus the share below 300, for the trait teacher minus the neutral teacher on the same questions; the expected first digit is the model's expected first digit of a new item, trait minus neutral, on shared neutral text. The eight animals of the main experiments were generated with 6,144 fresh questions per half, the other twenty with 64 questions answered 48 times, and the correlations hold within each group (weight +0.58 and +0.48, speed +0.67 and +0.50). Weight is a typical adult mass and speed the fastest mode of locomotion, a dive for the falcon and swimming for the whale and the dolphin.}
\label{sitab:11}
{\tabsize\setlength{\tabcolsep}{4pt}
\begin{tabular}{@{}lrrrr@{}}
\toprule
\textbf{Animal} & \textbf{Body weight} & \textbf{Top speed (km/h)} & \textbf{Numbers $\geq$ 700 minus < 300 (pp)} & \textbf{Expected first digit}\\
\midrule
lion & 190 kg & 80 & +8.1 & +0.137\\
elephant & 6 t & 40 & +5.5 & +0.069\\
bear & 300 kg & 50 & +3.2 & +0.070\\
cheetah & 50 kg & 110 & +2.8 & +0.073\\
giraffe & 1 t & 55 & +2.5 & +0.068\\
crow & 500 g & 50 & +2.5 & +0.047\\
whale & 50 t & 30 & +2.3 & +0.045\\
horse & 500 kg & 70 & +1.9 & +0.057\\
quoll & 1.5 kg & 15 & +1.4 & +0.024\\
chub & 1 kg & 8 & +0.9 & +0.015\\
warbler & 12 g & 30 & +0.8 & -0.022\\
loris & 1 kg & 2 & +0.1 & +0.009\\
falcon & 1 kg & 320 & -0.1 & +0.023\\
rabbit & 2 kg & 60 & -0.2 & +0.026\\
dolphin & 150 kg & 40 & -0.3 & +0.016\\
raven & 1.2 kg & 50 & -0.3 & +0.020\\
ant & 1 mg & 0.3 & -0.4 & -0.006\\
finch & 20 g & 30 & -0.4 & -0.007\\
hummingbird & 4 g & 50 & -0.6 & -0.028\\
snake & 5 kg & 15 & -0.8 & -0.004\\
slug & 10 g & 0.01 & -0.9 & -0.029\\
bee & 100 mg & 25 & -1.9 & +0.004\\
sloth & 5 kg & 0.2 & -2.1 & +0.011\\
owl & 1.5 kg & 65 & -2.4 & -0.019\\
turtle & 30 kg & 0.3 & -2.4 & +0.011\\
snail & 10 g & 0.05 & -2.7 & -0.032\\
tortoise & 100 kg & 0.3 & -2.8 & +0.017\\
mouse & 20 g & 13 & -3.2 & -0.029\\
\bottomrule
\end{tabular}}
\end{table*}

\end{document}